%% file: manuscript.tex
\documentclass{article}

    \PassOptionsToPackage{numbers, compress}{natbib}

\usepackage[preprint]{preprint_style}

\usepackage[utf8]{inputenc} 
\usepackage[T1]{fontenc}    
\usepackage{hyperref}       
\usepackage{url}            
\usepackage{fontawesome5}   
\usepackage{booktabs}       
\usepackage{amsfonts}       
\usepackage{nicefrac}       
\usepackage{microtype}      
\usepackage[table]{xcolor}  
\usepackage{amsmath}
\usepackage{amssymb}
\usepackage{bm}
\usepackage{graphicx}
\usepackage{multirow}
\usepackage{makecell}
\usepackage{adjustbox}
\usepackage{wrapfig}
\usepackage{caption}
\usepackage{listings}
\usepackage{algorithm}
\usepackage{algpseudocode}
\usepackage[most]{tcolorbox}

\lstdefinestyle{harnessprompt}{
  basicstyle=\ttfamily\scriptsize,
  showstringspaces=false,
  breaklines=true,
  breakatwhitespace=true,
  columns=fullflexible,
  keepspaces=true,
  frame=single,
  framerule=0.3pt,
  framesep=4pt,
  rulecolor=\color{black!30},
  backgroundcolor=\color{black!2},
  xleftmargin=4pt,
  xrightmargin=4pt,
}
\lstdefinestyle{harnessjson}{
  basicstyle=\ttfamily\scriptsize,
  showstringspaces=false,
  breaklines=true,
  breakatwhitespace=true,
  columns=fullflexible,
  keepspaces=true,
  frame=single,
  framerule=0.3pt,
  framesep=4pt,
  rulecolor=\color{black!30},
  backgroundcolor=\color{blue!2},
  xleftmargin=4pt,
  xrightmargin=4pt,
}

\newcommand{\cmark}{\textcolor{green!50!black}{\checkmark}}
\newcommand{\xmark}{\textcolor{red!70!black}{\ensuremath{\times}}}
\newcommand{\pmark}{\textcolor{orange!80!black}{\ensuremath{\triangle}}}
\definecolor{bestcell}{RGB}{191,230,203}
\definecolor{secondcell}{RGB}{255,248,220}
\newcommand{\best}[1]{\cellcolor{bestcell}\textbf{#1}}
\newcommand{\second}[1]{\cellcolor{secondcell}#1}
\title{DocAtlas: Long-Document Understanding as Mutable-State Interaction}

\author{%
  \quad \textbf{Hongchen Wei}$^{1,{\dagger},{\ddagger}}$
  \quad \textbf{Yuanzhe Wang}$^{2,{\dagger},{\ddagger}}$
  \quad \textbf{Bei Liu}$^{2,*}$ 
  \quad \textbf{Yifan Yang}$^{2}$ \\
  \quad \textbf{Qi Dai}$^{2}$
  \quad \textbf{Kai Qiu}$^{2}$
  \quad \textbf{Yunsheng Li}$^{2}$
  \quad \textbf{Dongdong Chen}$^{2}$ \\
  \quad \textbf{Chong Luo}$^{2}$
  \quad \textbf{Zhenzhong Chen}$^{1}$
  \quad \textbf{Baining Guo}$^{2}$ \\[2ex]
  $^1$Wuhan University \quad $^2$Microsoft 
}

\begin{document}

\maketitle
\begingroup
\renewcommand\thefootnote{}
\footnotetext{$^\dagger$ Equal contribution. $^\ddagger$ Work done during an internship at MSRA. $^*$ Project leader.}
\endgroup

\begin{abstract}
Long-document understanding requires models to find and combine evidence across many pages, layouts, tables, figures, and charts. Existing retrieval-augmented systems usually select evidence from a static index before generation, while recent agentic systems add multi-turn tool use but often rely on frozen proprietary backbones whose behavior is set by prompts. We present DocAtlas, a system that treats long-document understanding as a mutable-state information-seeking process. We instantiate DocAtlas as a mutable document harness: an external environment that determines what document information is searched, read, stored, reviewed, and shown to the model at each step. Given a document and question, the harness exposes search, reading, note-taking, and review tools, maintains a hierarchical tree and note store, and updates both as the agent records evidence. DocAtlas combines self-improving retrieval, selective evidence access, and active working memory under a fixed context budget. The same harness supports inference-time use with large VLMs and end-to-end reinforcement learning for compact VLM agents. With GPT-5.4, DocAtlas reaches 71.4\% on MMLongBench-Doc, exceeding the human-expert reference of 65.8\%. A Qwen3.5-4B VLM trained with end-to-end RL in the DocAtlas environment reaches 63.7\%, compared with a 54.4\% direct-input baseline, showing that mutable document-harness design can improve compact document agents by a large margin.
Project homepage: \href{https://officeintelligence.github.io/docatlas/}{\faGlobe}.
\end{abstract}

\section{Introduction}

Real-world documents such as financial reports, legal contracts, scientific papers, and government filings often spread important information across dozens or even hundreds of pages. They also combine free text with tables, figures, and charts in varied layouts. Answering natural-language questions over these documents therefore requires finding, extracting, and combining evidence that is distributed across pages and modalities \cite{ding2025survey}. This remains difficult because relevant evidence may appear anywhere in a long document, visual elements such as charts and tables cannot be handled by text extraction alone \cite{cho2024m3docrag}, and complex questions often require evidence from multiple distant regions or modalities \cite{ma2024mmlongbench}. Simply feeding the full document into a model quickly runs into the practical limits of current vision-language models \cite{hu2025mplug,liu2026textmonkey,ye2023mplug}. Even when the input fits within the context window, performance often drops because irrelevant content competes for attention \cite{liu2024lost}. These issues call for systems that can navigate long documents selectively, ground answers in both text and visual content, and combine information across distant parts of a document.

\begin{figure*}[t]
\centering
\begin{minipage}[t]{0.55\textwidth}
\vspace{0pt}
\centering
\scriptsize
\setlength\tabcolsep{3.5pt}
\renewcommand{\arraystretch}{1.08}
\begin{tabular}{lcccc}
\toprule
\textbf{Method} & \makecell{\textbf{Mutable}\\\textbf{state}} & \makecell{\textbf{Flexible}\\\textbf{interaction}} & \makecell{\textbf{Grounded}\\\textbf{memory}} & \makecell{\textbf{Trainable}\\\textbf{RL policy}} \\
\midrule
M3DocRAG\cite{cho2024m3docrag} & \xmark & \xmark & \xmark & \xmark \\
DocAgent\cite{yang2025docagent} & \xmark & \pmark & \xmark & \xmark \\
SimpleDoc\cite{jain2025simpledoc} & \xmark & \pmark & \pmark & \xmark \\
DocLens\cite{zhu2025doclens} & \xmark & \cmark & \cmark & \xmark \\
DocDancer\cite{zhang2026docdancer} & \pmark & \cmark & \xmark & \pmark \\
MACT\cite{yu2025mact} & \pmark & \pmark & \xmark & \xmark \\
DocAtlas & \cmark & \cmark & \cmark & \cmark \\
\bottomrule
\end{tabular}

\vspace{8pt}
{\footnotesize (a) Design comparison. \cmark: yes, \pmark: partial, \xmark: no.}
\end{minipage}
\hfill
\begin{minipage}[t]{0.42\textwidth}
\vspace{0pt}
\centering
\includegraphics[width=\linewidth]{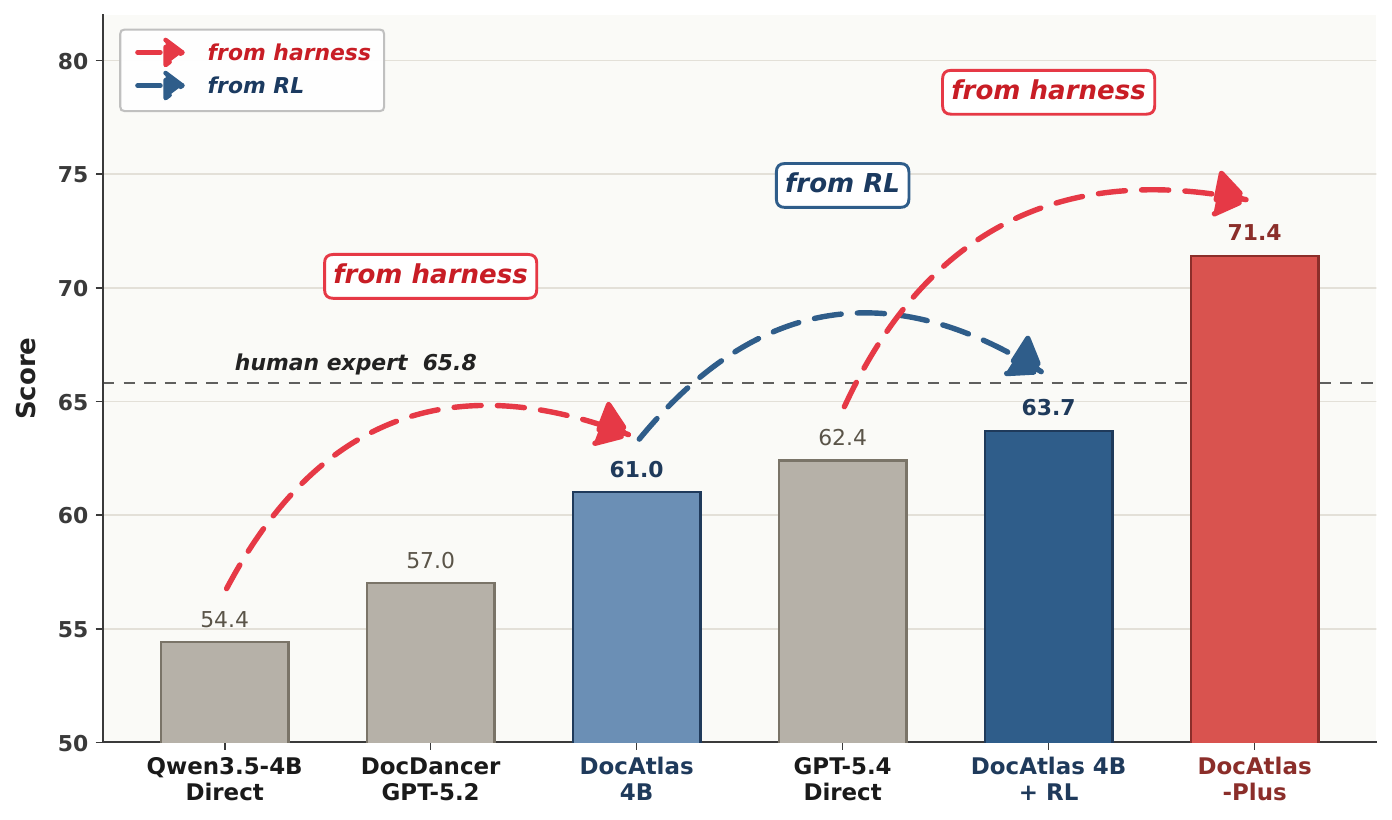}
\vspace{-4pt}
{\footnotesize (b) MMLongBench-Doc ALL.}
\end{minipage}
\caption{Motivation and positioning of DocAtlas. (a) \cmark, \pmark, and \xmark denote full, partial, and absent support, respectively. Flexible interaction means that the agent can choose the order and arguments of tool calls rather than following a fixed pipeline; grounded memory means that source-attributed evidence can be stored and queried in later steps. DocAtlas combines mutable state, flexible interaction, grounded memory, and a trainable compact policy. (b) On MMLongBench-Doc, DocAtlas-Plus (our GPT-5.4 instantiation) exceeds the human-expert reference, while the same environment improves Qwen3.5-4B \cite{Qwen35} from direct input to DocAtlas 4B and then to DocAtlas 4B+RL.}
\label{fig:motivation}
\end{figure*}

Early work \cite{cuconasu2024power,gao2023retrieval} has addressed part of this problem with retrieval-augmented generation (RAG). 
In these systems, an embedding-based retriever selects a fixed set of relevant pages, and a vision-language model (VLM) generates an answer in one pass \cite{cho2024m3docrag, han2025mdocagent}. This design works reasonably well for simple lookups, but it gives the model no control over which pages to inspect more carefully or how to revise the search based on partial findings. To overcome this limitation, recent work has moved toward agentic methods, where a vision-language model interacts with a document through iterative tool use for searching, reading, and reasoning \cite{jain2025simpledoc,sun2025docagent,wu2025docreact,yao2023react,yu2025mact,zhu2025doclens}. However, these systems still depend on frozen proprietary backbones, and their agent behavior is largely specified through prompting rather than learned from data. 
A natural next question is whether the agent itself can be trained. 
DocDancer \cite{zhang2026docdancer} takes a simpler approach by training a single open-source model on synthesized trajectories with supervised fine-tuning. 
However, DocDancer uses a text-only LLM as the controller, and its \texttt{Read} tool relies on an external VLM for visual understanding. 
As a result, the agent learns when to call a tool, but not how to interpret visual content itself. 
Because it is trained by imitation, it is also limited by the coverage of the expert trajectories it observes.

DocAtlas addresses these gaps by exposing the document interface as a mutable environment in which the agent policy can be trained. 
Equivalently, DocAtlas can be viewed as a mutable document harness: it wraps a VLM with tools and state that determine what document information is searched, read, stored, reviewed, and shown at each step. The agent does more than call tools over a fixed document representation. Each episode keeps an environment state with the document, a hierarchical tree, a structured note store, and the set of explored pages. \textsc{Search} reads this updated state; \textsc{Read} exposes selected pages as markdown, crops, and page images; \textsc{Note} records source-attributed findings and writes them back to the tree; and \textsc{Review} revisits prior notes when later reasoning requires them. This loop makes later retrieval depend on earlier evidence and helps the agent reason across many steps under a fixed context budget. Because the same interaction protocol is used at inference time and during RL, DocAtlas also provides a direct way to train compact VLM agents instead of treating tool use as a fixed prompt script. With GPT-5.4 \cite{gpt54}, DocAtlas reaches 71.4\% on MMLongBench-Doc \cite{ma2024mmlongbench}, surpassing the human-expert reference at 65.8\%. A Qwen3.5-4B \cite{Qwen35} VLM fine-tuned with end-to-end RL in the same environment reaches 63.7\%, compared with a 54.4\% direct-input baseline.

This work makes three contributions:
\begin{itemize}
    \item We formulate long-document understanding as a mutable-state information-seeking process in which reading and note-taking update the retrieval and memory state used by decisions.
    \item We instantiate this formulation in DocAtlas, a harness that combines self-improving retrieval, decoupled search and reading, and structured note and review operations.
    \item We show that the same environment supports both large VLM agents at inference time and end-to-end RL for compact VLMs: DocAtlas reaches 71.4\% on MMLongBench-Doc with GPT-5.4, while an RL-tuned Qwen3.5-4B reaches 63.7\% against a 54.4\% direct-input baseline.
\end{itemize}

\section{Related Work}

\textbf{Multimodal Retrieval for Long-Document Understanding.}
\label{sec:rag}
Multi-page document QA \cite{gong2025mhier,napolitano2024leveraging,xiong2026docr1,zheng2026doc} is commonly handled with multimodal RAG: pages are embedded as visual or textual vectors, a fixed top-$k$ subset is retrieved, and a VLM answers in one pass. Recent visual retrievers and benchmarks, including ColPali \cite{FaysseSWOVHC25}, VisRAG \cite{yu2024visrag}, ViDoRe \cite{mace2025vidore}, and MIRACL-VISION \cite{osmulski2025miraclvision}, show the value of page-image retrieval for documents with complex layouts and visual content. M3DocRAG \cite{cho2024m3docrag} and MDocAgent \cite{han2025mdocagent} build long-document QA pipelines on this static-index design with visual or parallel text--image retrieval. Adaptive retrieval has also been studied in text QA through query rewriting \cite{ma2023query} and self-reflective retrieval control \cite{asai2024selfrag}. These methods adapt queries or retrieval decisions, but the document index itself remains precomputed. DocAtlas instead updates the retrieval state within an episode by writing page-grounded findings back into a hierarchical tree.

\textbf{Agentic Document Understanding.}
\label{sec:agentic}
Recent systems \cite{chen2025hear,gomes2025visdocsketcher,li2026deepread,liu2025resolving} move from one-shot retrieval to tool-mediated interaction: Doc-React iterates sub-queries \cite{wu2025docreact}, SimpleDoc combines visual embeddings with page summaries \cite{jain2025simpledoc}, and DocLens separates page navigation from element localization \cite{zhu2025doclens}. Closest to our tree-based navigation, DocAgent \cite{sun2025docagent} builds a structured XML outline with section hierarchy, page ranges, paragraph hints, captions, and identifiers for fetching full content. Its outline is primarily a fixed navigation scaffold, whereas DocAtlas treats the tree as per-question mutable state enriched by \textsc{Note} and observed by later \textsc{Search} calls. 
Existing document agents can be viewed as hand-designed harnesses that decide what information is retrieved, shown, and carried forward, consistent with recent LLM harness work \cite{lee2026meta,lou2026autoharness}. However, they typically rely on frozen proprietary backbones, follow staged tool flows, or leave the document state unchanged. Learning-based systems such as DocDancer \cite{zhang2026docdancer} train tool-use behavior, but either spread credit across multiple agents or rely on an external VLM for visual reading. DocAtlas puts visual reading, control, mutable memory, and RL optimization in a single VLM policy.

\section{Method}
\label{sec:method}

Passing an entire long document to a VLM quickly exceeds the effective context window. Even when the document fits, irrelevant content can dilute attention and hurt performance \cite{liu2024lost}. Static retrieval selects a fixed subset of pages, but it cannot refine the search as evidence accumulates. We instantiate DocAtlas as a mutable document harness. Given a document and question, the harness exposes a set of tools, maintains a document tree and note store, and updates both as the agent searches, reads, and records evidence. Unlike a static retrieval harness, the state seen by later tool calls depends on earlier interactions. We formalize this as a mutable-state information-seeking process: the agent's actions reveal document content and update the retrieval and memory state used by later steps.

\textbf{Terminology.} We use \emph{model} or \emph{backbone} for the underlying VLM weights, \emph{policy} for the mapping from interaction history to the next action, and \emph{agent} for the policy running inside the DocAtlas tool environment. Thus, the same VLM can be described as a model when discussing its architecture or size, as a policy when discussing RL or action choices, and as an agent when it interacts with tools and environment state.

Given a document $\mathcal{D} = \{d_1, \ldots, d_N\}$ consisting of $N$ pages and a natural-language query $q$, the agent produces a trajectory
\begin{equation}
\tau = (q,\; u_1, o_1,\; u_2, o_2,\; \ldots,\; u_T, a),
\label{eq:trajectory}
\end{equation}
where each action $u_t$ for $t < T$ is a tool call $(\texttt{tool}_t, \texttt{arg}_t) \in \mathcal{T} \times \mathcal{A}$, and the trajectory ends with a terminal action $u_T = \texttt{finish}(a)$ that outputs the final answer $a$. For each tool call, the environment returns an observation $o_t = \texttt{env}(\texttt{tool}_t, \texttt{arg}_t;\, \mathcal{S}^{(t)})$, where $\mathcal{S}^{(t)} = (\mathcal{D}, \mathcal{G}^{(t)}, \mathcal{M}^{(t)}, \mathcal{H}^{(t)})$ is the environment state comprising the document $\mathcal{D}$, the document index $\mathcal{G}^{(t)}$ (defined in \S\ref{sec:adaptive_index}), the note store $\mathcal{M}^{(t)}$ (\S\ref{sec:memory}), and the set of previously explored pages $\mathcal{H}^{(t)} \subseteq [1, N]$. The tool set $\mathcal{T} = \{\textsc{Search}, \textsc{Read}, \textsc{Note}, \textsc{Review}\}$ has a fixed action space but no fixed execution order: unlike staged pipelines, the agent may invoke any tool at any step and interleave search, reading, and memory operations as needed.

The key departure from static retrieval is that $\mathcal{S}^{(t)}$ is mutable. In particular, $\mathcal{G}^{(t)}$ and $\mathcal{M}^{(t)}$ are updated by the agent's own actions: \textsc{Note} writes evidence-grounded annotations into the tree and archives structured notes, while \textsc{Review} exposes selected notes back to the policy. Thus, a later \textsc{Search} call depends on the original document, the query, and what the agent has already read and recorded. This makes retrieval part of the closed-loop decision process rather than a fixed preprocessing step.

\begin{figure}[t]
    \centering
    \includegraphics[width=\linewidth]{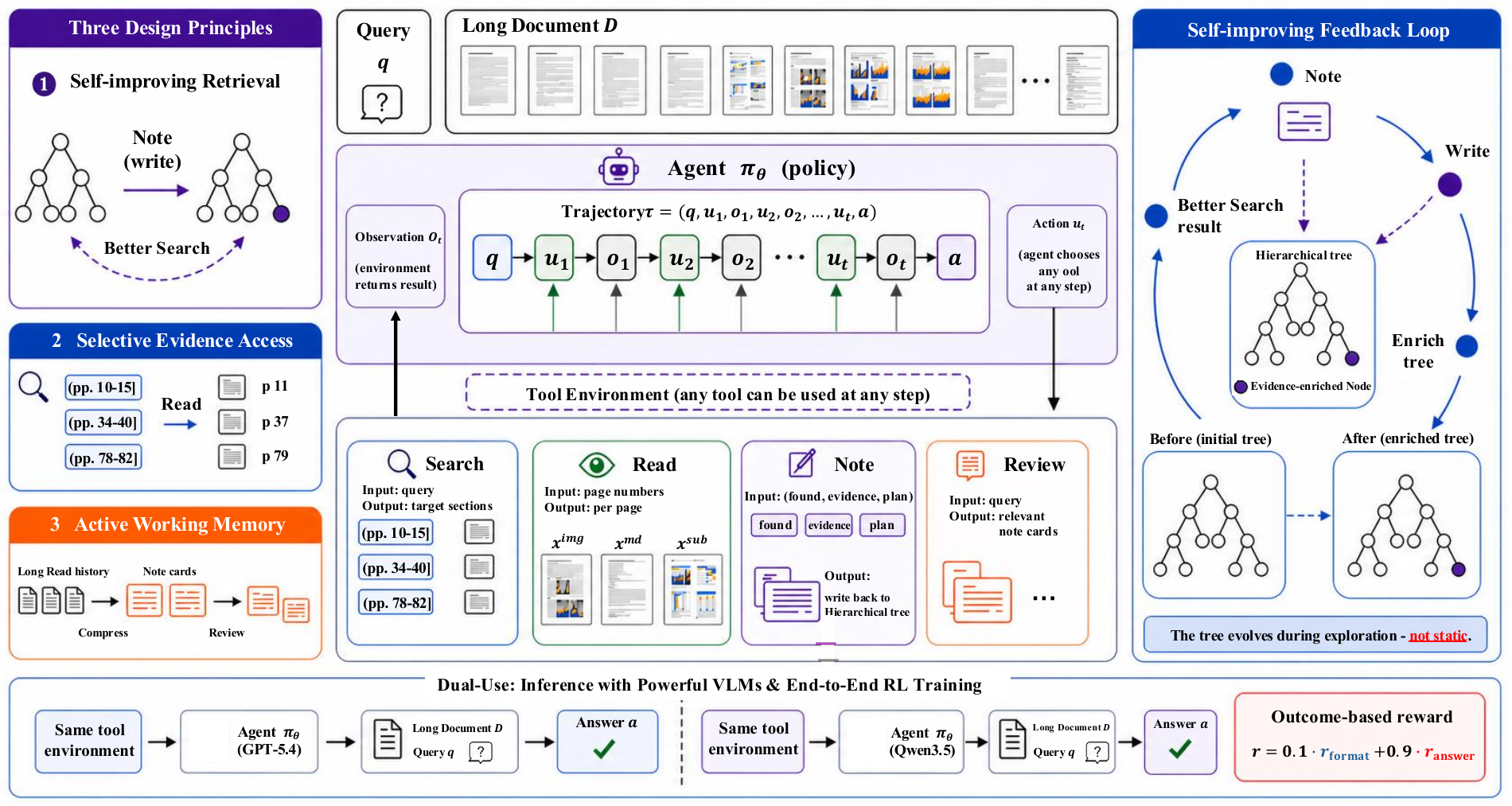}
    \caption{Overview of DocAtlas. The agent searches a mutable tree, reads selected document pages in multimodal form, writes structured notes that both compress context and update the retrieval state, reviews prior findings when needed, and finally produces an answer. Unlike static RAG, evidence gathered during reading changes the state used by later search and reasoning steps.}
    \label{fig:overview}
\end{figure}

DocAtlas is built around a tool environment with three design principles (Figure~\ref{fig:overview}). Self-improving retrieval (\S\ref{sec:adaptive_index}) organizes the document as a hierarchical tree that \textsc{Search} can navigate and update during exploration. Selective evidence access (\S\ref{sec:decoupled}) separates finding evidence from consuming it: \textsc{Search} identifies relevant sections, and \textsc{Read} lets the agent decide which pages to inspect and in which modality. Active working memory (\S\ref{sec:memory}) is implemented through \textsc{Note} and \textsc{Review}, which record, optionally archive, and review evidence so the agent can reason across many steps under a fixed context budget. This interaction can be viewed as a sequential decision problem, which we use in \S\ref{sec:rl} for end-to-end reinforcement learning.

\subsection{Self-Improving Retrieval}
\label{sec:adaptive_index}

Standard embedding-based retrieval maps each page to a fixed-dimensional vector. 
This discards the hierarchical structure of long documents and yields a static index that cannot improve as the agent gathers evidence. 
Inspired by PageIndex\footnote{\url{https://github.com/VectifyAI/PageIndex}}, which builds a hierarchical tree index and performs LLM-guided tree search, DocAtlas uses a tree-structured index for semantic navigation. 
We extend this idea in two ways: tree construction incorporates visual parsing for multimodal document understanding, and \textsc{Search} is decoupled from \textsc{Read} so retrieval proposes candidate regions rather than directly determining the consumed evidence.

\textbf{Hierarchical index.}
Each document $\mathcal{D}$ is organized offline into a tree $\mathcal{G} = (V, E)$, where each node $v \in V$ stores $(title_v,\; [p^v_s, p^v_e],\; summary_v,\; findings_v)$, namely a section title, page range, summary, and a mutable findings list (initially empty). 
The tree is constructed once per document by a VLM in a question-agnostic manner: the model receives structured markdown extracted from the document together with parsed visual content, including figure/table/chart captions and layout-derived cues, and outputs a JSON tree with section titles, page ranges, and summaries. 
This preprocessing does not use benchmark questions, answers, or evidence annotations. 

\textbf{Structure-aware navigation.}
\textsc{Search} takes a search query $q_s$ (which may differ from $q$, e.g., a sub-question) and the current index state:
\begin{equation}
\textsc{Search}(q_s;\, \mathcal{G}^{(t)}) \;\to\; \bigl\{(v_1, s_1),\, \ldots,\, (v_k, s_k)\bigr\},
\label{eq:search}
\end{equation}
An auxiliary LLM receives the serialized tree, including titles, page ranges, summaries, and accumulated annotations, together with $q_s$ and explored pages $\mathcal{H}^{(t)}$. It then selects tree nodes by using the hierarchy, preferring unvisited regions when useful, and using annotations from prior exploration.

\textbf{Evidence-grounded enrichment.}
When the agent records findings via \textsc{Note}, the evidence entries $\texttt{evidence}_t \subseteq \{(e_j, p_j, \text{type}_j)\}$ (Eq.~\ref{eq:note}) are propagated back to the tree. For each cited page $p$, the finest-grained covering node $v^* = \arg\min_{v:\, p \in [p^v_s, p^v_e]} |p^v_e - p^v_s|$ receives a compact annotation in $findings_{v^*}$. The index thus evolves:
\begin{equation}
\mathcal{G}^{(t+1)} = \texttt{enrich}\bigl(\mathcal{G}^{(t)},\; \texttt{evidence}_t\bigr),
\label{eq:enrich}
\end{equation}
so later \textsc{Search} calls see accumulated findings alongside original summaries. Each annotation contains a normalized finding, its source page, evidence type, and the step at which it was observed. The enrichment operation is deliberately conservative: it appends or merges source-attributed findings into the finest covering node but never rewrites the original section summary. Annotations are navigation hints only and do not replace direct evidence when composing the final answer.

\subsection{Selective Evidence Access}
\label{sec:decoupled}

In most existing agentic systems, retrieval and reading are tightly coupled: if search returns a candidate set $C = \{v_1, \ldots, v_k\}$, all pages in $\texttt{pages}(C) \triangleq \bigcup_{v \in C} [p^v_s, p^v_e]$ are consumed. This wastes context when a relevant section spans many pages and prevents accessing pages outside $C$ when evidence is referenced indirectly. DocAtlas decouples these stages: \textsc{Search} proposes candidate regions, while \textsc{Read} gives the agent explicit control over the final evidence set $P \subseteq [1, N]$. Retrieval results are therefore proposals rather than mandatory context; the agent may read pages from a search hit, neighboring pages, explicit page references, or unresolved gaps recorded in memory.

\textbf{Multimodal reading.}
The agent selects a page subset $P$ and invokes:
\begin{equation}
\textsc{Read}(P) \;\to\; \bigl[(x^{\text{img}}_p,\; x^{\text{md}}_p,\; x^{\text{sub}}_p)\bigr]_{p \in P},
\label{eq:read}
\end{equation}
returning, for each page $p$, three content types: a full-page layout image $x^{\text{img}}_p$, a structured markdown rendering $x^{\text{md}}_p$ (extracted by MinerU \cite{niu2025mineru2}), and a set of cropped sub-images $x^{\text{sub}}_p = \{x^{\text{sub}}_{p,1}, \ldots\}$ for figures and charts. 
The read observation is assembled adaptively rather than fixed by backbone type. 
By default, the agent receives the page layout image together with markdown, so text, layout, and global spatial context are jointly available. 
For questions that require fine-grained visual inspection, \textsc{Read} further attaches cropped sub-images to zoom in on the relevant regions. 
This makes \textsc{Read} the main channel through which visual observations enter the agent and the tool whose use can be improved by outcome-based RL.

\subsection{Active Working Memory}
\label{sec:memory}

Each \textsc{Read} invocation appends multimodal observations to the interaction history $h_t$. 
At step $t$, the cumulative token cost $\sum_{i \leq t}(|u_i| + |o_i|)$ grows toward the context budget $B$, forcing a tradeoff between retaining early evidence and allocating tokens to further exploration. 
DocAtlas therefore separates active context from evidence memory: bulky page observations remain in the interaction history when available, while \textsc{Note} stores compact, source-attributed findings that can be retrieved later by \textsc{Review}. 

\textbf{Structured note-taking.}
\textsc{Note} records findings as a structured tuple
\begin{equation}
n = \bigl(\texttt{found},\;\; \texttt{evidence},\;\; \texttt{plan}\bigr),
\label{eq:note}
\end{equation}
where \texttt{found} summarizes what has been established, \texttt{evidence} $= \{(e_j, p_j, \text{type}_j)\}_{j=1}^{J}$ is an array of typed, source-attributed entries ($\text{type}_j \in \{\text{text}, \text{table}, \text{image}, \text{formula}\}$), and \texttt{plan} states remaining information gaps. Unlike generic transcript summarization, notes in DocAtlas are structured evidence objects: they store what has been established, where it was observed, what modality supports it, and what remains unresolved. \textsc{Note} triggers two side effects. When archival is enabled, prior \textsc{Read} outputs in $h_t$ can be replaced in-place with short placeholders, yielding $\hat{h}_t$ with $|\hat{h}_t| \ll |h_t|$. Evidence entries are also written back to the tree (Eq.~\ref{eq:enrich}), closing the feedback loop with \textsc{Search}.

To avoid turning memory into an unconstrained summary channel, notes are constrained to be extractive and page-grounded: each evidence entry must be copied or normalized from a previously read page and paired with its source page and modality. Tree annotations therefore act as source-attributed navigation hints rather than free-form model beliefs. We use append-and-merge semantics: new entries are concatenated with existing ones rather than replacing them, so the auxiliary model can inspect the accumulated evidence trail during later search.

\textbf{Evidence review.}
As notes accumulate in $\mathcal{M}^{(t)}$, \textsc{Review}$(q_v)$ reviews the note store and returns only the findings relevant to a query $q_v$. 
Each note $n_i \in \mathcal{M}^{(t)}$ is projected to a compact note card $c_i = (\texttt{id}_i, \texttt{step}_i, \texttt{found}_i, \texttt{pages}_i)$ that strips the full evidence payload. 
An auxiliary LLM receives $\{c_1, \ldots, c_{|\mathcal{M}^{(t)}|}\}$ and selects the relevant subset (greedy decoding; prompt in Appendix~\ref{app:prompts}). 
\textsc{Review} exposes source-attributed evidence summaries to the policy without requiring the full multimodal transcript to remain in context.

\subsection{End-to-End RL over Agent Trajectories}
\label{sec:rl}

Because DocAtlas exposes document understanding as a sequence of structured actions and observations, it can train compact VLM agents directly. 
We optimize a single VLM policy over complete tool trajectories using outcome rewards. 
The environment components, including the tree, tool executors, and auxiliary subroutines used by \textsc{Search} and \textsc{Review}, are frozen; 
the policy learns when to search, what to read, what to record, and when to answer. 

\textbf{Formulation.}
The agent policy $\pi_\theta$ is a single VLM that, at each step $t$, conditions on $h_t = (q, u_1, o_1, \ldots, u_{t-1}, o_{t-1})$ and emits $u_t \sim \pi_\theta(\cdot \mid h_t)$. 
During RL training, each action is either a structured tool call or a terminal answer normalized as \texttt{\textbackslash boxed\{\}} for reward parsing; 
the trajectory terminates at the answer or after $T_{\max}{=}8$ turns. 
The four tools, the tree, and the auxiliary LLM subroutines inside \textsc{Search} and \textsc{Review} constitute the frozen environment; they receive no gradient updates. 

\textbf{Reward and optimization.}
RL rewards are computed from the final answer only. If the last visible answer lacks a parsable \texttt{\textbackslash boxed\{\}} field, the reward is zero; otherwise, we extract the boxed content and apply the LongDocURL type-aware answer score:
\begin{equation}
r = \mathrm{Score}_{\mathrm{LongDocURL}}\bigl(\mathrm{ExtractBoxed}(a), a^\star\bigr).
\label{eq:reward}
\end{equation}
We optimize with GRPO \cite{shao2024deepseekmath} using DAPO-style asymmetric clipping \cite{yu2025dapo} ($\epsilon_{\text{low}}{=}0.1$, $\epsilon_{\text{high}}{=}0.3$), $n{=}8$ rollouts, group-normalized advantages, and token-level KL regularization ($\lambda_{\text{KL}}{=}0.01$). Appendix~\ref{app:training} gives full reward and training details.

\section{Experiments}
\label{sec:experiments}

We organize the evaluation around two questions: 
\textbf{Q1:} Does a mutable tree-based tool environment improve long-document understanding on benchmarks with complex layouts and visual content? 
\textbf{Q2:} Can the same environment train compact VLM policies through end-to-end RL, instead of only serving as an inference-time prompting scaffold? 
We report results for large VLM agents at inference time and RL-trained Qwen3.5 policies.

\subsection{Experimental Setup}
\label{sec:exp_setup}
\begin{table*}[t]
\centering
\caption{Main results on MMLongBench-Doc \cite{ma2024mmlongbench}, FinRAGBench-V \cite{zhao2025finragbench}, and LongDocURL \cite{deng2025longdocurl}. MMLongBench-Doc is reported by evidence source: text (TXT), layout (LAY), chart (CHA), table (TAB), figure (FIG), unanswerable (UNA), with overall accuracy (Acc), F1, and LLM-as-judge score (LasJ). FinRAGBench-V reports LLM-as-judge scores by evidence type and overall. LongDocURL reports overall LLM-as-judge score using GPT-5.4. Compact DocAtlas rows use GPT-5.4 as a fixed reference auxiliary for \textsc{Search}/\textsc{Review}; Table~\ref{tab:aux_sensitivity} tests open-weight and self auxiliaries. DocLens and $^{\dag}$ rows are copied from prior papers and are included for comparison; unavailable entries are marked with --. Qwen RL rows omit LongDocURL because LongDocURL is used for RL data construction. Green cells mark the best-performing method in each metric column, and pale yellow cells mark the second-best-performing method; $^{*}$ denotes results surpassing the MMLongBench-Doc human-expert.}
\scriptsize
\renewcommand{\arraystretch}{1.08}
\setlength\tabcolsep{2.5pt}
\begin{tabular}{@{}l ccccccccc|cccc|c@{}}
\toprule
\multirow{2}{*}{\textbf{Model}} & \multicolumn{9}{c}{MMLongBench-Doc} & \multicolumn{4}{c}{FinRAGBench-V} & LongDocURL \\
\cmidrule(r){2-10} \cmidrule(lr){11-14} \cmidrule(l){15-15}
& \textbf{TXT} & \textbf{LAY} & \textbf{CHA} & \textbf{TAB} & \textbf{FIG} & \textbf{UNA} & \textbf{Acc} & \textbf{F1} & \textbf{LasJ} & \textbf{TXT} & \textbf{TAB} & \textbf{CHA} & \textbf{LasJ} & \textbf{LasJ} \\
\midrule
\multicolumn{15}{@{}c}{\textit{Vanilla VLMs}} \\ \midrule
GPT-5.2               & 45.8 & 45.6 & 45.3 & 43.3 & 33.5 & \best{86.0} & 52.8 & 52.1 & 57.6   & 40.4 & 28.9 & 40.7 & 36.0 & 59.4 \\
GPT-5.4               & 57.8 & 57.7 & 55.7 & 61.1 & 51.7 & \second{75.6} & 62.4 & 59.4 & 63.4   & 64.6 & 45.2 & 59.9 & 55.1 & 66.9 \\
Claude-4-Sonnet       & 50.4 & 49.4 & 50.5 & 57.3 & 43.9 & 59.0 & 53.4 & --   & --   & 36.6 & 20.2 & 51.9 & 33.8 & -- \\
Gemini-2.5-Flash      & 44.0 & 53.2 & 46.0 & 43.9 & 48.2 & 56.7 & 49.6 & --   & --   & 49.0 & 41.6 & 41.0 & 43.0 & -- \\
Gemini-2.5-Pro        & 52.1 & 62.1 & 55.5 & 55.3 & 54.0 & 59.9 & 58.1 & --   & --   & 62.2 & 55.3 & 50.4 & 54.9 & -- \\
Qwen3.5-4B            & 47.5 & 48.7 & 47.6 & 55.4 & 45.5 & 63.4 & 54.4 & 53.1 & 58.7   & 66.2 & 49.0 & 48.8 & 52.8 & 52.4 \\
Qwen3.5-9B            & 54.2 & 53.8 & 51.0 & 55.7 & 46.4 & 66.9 & 58.0 & 55.0 & 60.2   & 64.2 & 52.2 & 54.7 & 55.8 & 55.5 \\
Qwen3.5-397B-A13B     & --   & --   & --   & --   & --   & --   & 61.9 & --   & --   & -- & -- & -- & --   & -- \\
\midrule
\multicolumn{15}{@{}c}{\textit{VLMs Augmented with OCR}} \\ \midrule
Claude-4-Sonnet       & 52.7 & 51.6 & 50.0 & 58.1 & 45.3 & 65.9 & 56.0 & --   & --   & 58.7 & 21.6 & 54.3 & 41.0 & -- \\
Gemini-2.5-Flash      & 55.9 & 54.9 & 52.7 & 63.4 & 50.3 & 60.8 & 58.5 & --   & --   & 67.6 & 64.4 & 46.1 & 58.3 & -- \\
Gemini-2.5-Pro        & 59.7 & 65.3 & 60.8 & 68.3 & 55.7 & 58.4 & 63.3 & --   & --   & 70.0 & 70.0 & 56.2 & 64.9 & -- \\
\midrule
\multicolumn{15}{@{}c}{\textit{VLM-based Agentic Frameworks}} \\ \midrule
M3DocRAG (w/ Qwen2-VL-7B)$^{\dag}$  & 30.0 & 23.5 & 18.9 & 20.1 & 20.8 &  5.8 & 21.0 & --   & --   & -- & -- & -- & --   & -- \\
MDocAgent (w/ GPT-4o)$^{\dag}$      & --   & --   & --   & --   & --   & --   & 42.0 & --   & --   & -- & -- & -- & --   & -- \\
DocDancer (w/ GPT-5.2)              & --   & --   & --   & --   & --   & --   & 57.0 & --   & --   & -- & -- & -- & --   & -- \\
\multicolumn{15}{@{}l}{\textbf{SimpleDoc}} \\
\quad w/ Claude-4-Sonnet      & 52.1 & 53.3 & 58.3 & 62.4 & 46.9 & 66.5 & 58.6        & --   & --   & 59.6 & 68.9 & 54.9 & 61.7 & -- \\
\quad w/ Gemini-2.5-Flash     & 45.5 & 57.4 & 49.0 & 51.6 & 45.2 & 66.5 & 53.3        & --   & --   & 70.2 & 56.2 & 53.6 & 58.3 & -- \\
\quad w/ Gemini-2.5-Pro       & 48.4 & 54.8 & 55.7 & 56.1 & 52.5 & 59.7 & 56.6        & --   & --   & 67.5 & 64.0 & 60.9 & 63.6 & -- \\
\multicolumn{15}{@{}l}{\textbf{DocLens}} \\
\quad w/ Claude-4-Sonnet      & 59.9 & 58.2 & 54.4 & 63.9 & 55.3 & 74.0 & 63.3        & --   & --   & 70.2 & 66.0 & 60.3 & 64.8 & -- \\
\quad w/ Gemini-2.5-Flash     & 59.5 & 61.5 & 54.8 & 66.9 & 59.0 & 73.8 & 64.7        & --   & --   & 69.9 & 71.3 & 64.5 & 68.5 & -- \\
\quad w/ Gemini-2.5-Pro       & 63.7 & 64.6 & 64.3 & 69.7 & 60.2 & 72.2 & 67.6$^{*}$  & --   & --   & 68.9 & 74.2 & 67.1 & 70.4 & -- \\
\midrule
\multicolumn{15}{@{}l}{\cellcolor{gray!15}\textbf{DocAtlas (Ours)}} \\
\quad w/ Qwen3.5-4B           & 58.4 & 54.6 & 56.1 & 67.0 & 48.1 & 70.6 & 61.0        & 58.7   & 63.5  & 73.3 & 66.5 & 65.4 & 67.9 & 72.5 \\
\quad w/ Qwen3.5-9B           & 58.8 & 55.3 & 56.2 & 67.8 & 51.9 & 70.8   & 61.6        & 59.4   & 64.7   & 73.9 & 67.1 & 66.4 & 69.8 & 74.0 \\
\quad RL w/ Qwen3.5-4B        & 59.2 & 55.9 & 57.5 & 67.8 & 51.8 & 74.6   & 63.7        & 62.4   & 67.9   & 75.2 & 70.7 & 68.5 & 71.7 & -- \\
\quad RL w/ Qwen3.5-9B        & 66.6 & 60.5 & 58.9 & 70.1 & 59.9 & 59.3   & 64.4        & 63.1   & 69.6   & 75.8 & 72.0 & 70.7 & 72.6 & -- \\
\quad w/ GPT-5.2              & \second{67.0} & \best{71.5} & \second{68.0} & \best{74.4} & \second{67.6} & 66.5 & \second{70.6$^{*}$}  & \second{69.6} & \second{73.9}   & \second{76.0} & \second{76.2} & \second{73.1} & \second{75.2} & \second{77.5} \\
\quad w/ GPT-5.4              & \best{68.8} & \second{69.0} & \best{68.2} & \second{74.1} & \best{69.8} & 68.6 & \best{71.4$^{*}$}  & \best{70.3} & \best{74.6}   & \best{76.3} & \best{77.4} & \best{73.2} & \best{75.6} & \best{78.8} \\
\midrule
Human Expert$^{\dag}$ & --   & --   & --   & --   & --   & --   & 65.8 & --   & --   & -- & -- & -- & --   & -- \\
\bottomrule
\end{tabular}
\label{tab:main_results}
\end{table*}

\paragraph{Benchmarks and metrics.}
We evaluate on three widely used long-document understanding benchmarks and follow their standard evaluation protocols. MMLongBench-Doc \cite{ma2024mmlongbench} reports overall accuracy, F1, LLM-as-judge score, and accuracy by evidence source. FinRAGBench-V \cite{zhao2025finragbench} reports LLM-as-judge scores overall and by evidence type. LongDocURL \cite{deng2025longdocurl} reports the overall LLM-as-judge score; we use GPT-5.4 as the judge. Since LongDocURL is used to construct the RL training set (\S\ref{sec:rl_data}), we do not report Qwen RL results on LongDocURL.

\paragraph{DocAtlas configuration.}
The environment exposes four tools, \textsc{Search}, \textsc{Read}, \textsc{Note}, and \textsc{Review}, with no fixed execution order. \textsc{Read} returns structured markdown, layout images, and cropped sub-images. \textsc{Note} records extractive, page-grounded evidence and optionally archives bulky read observations from the active context. The auxiliary LLMs inside \textsc{Search} and \textsc{Review} are frozen routing and memory-selection modules that can be swapped without changing the policy or tools; the main table uses GPT-5.4 as a fixed reference, and \S\ref{sec:analysis} tests open-weight and self auxiliaries. Archival is enabled for RL-trained compact policies because unbounded accumulation of tool observations makes multi-rollout training unstable (\S\ref{sec:rl_data}).

\paragraph{RL Data Construction and Training} 
\label{sec:rl_data}
We construct RL data from LongDocURL using model-specific pass@16 filtering. For each compact VLM policy, we sample 16 trajectories per question in the DocAtlas environment and retain medium-difficulty questions with at least one but not all correct rollouts; all correct questions provide little advantage signal, while all-wrong questions provide no positive trajectory under sparse rewards. Because filtering is performed separately for each policy, the retained sets differ: 612 questions for Qwen3.5-4B and 587 for Qwen3.5-9B. 
In early RL runs, the multimodal output of \textsc{Read} caused highly variable and rapidly growing contexts across rollouts, making memory use unstable. We therefore enable context archival during RL training: after the agent records a page-grounded note, bulky read observations can be replaced by short placeholders while the structured evidence remains accessible through \textsc{Review}. This reduces model input length and makes training more stable. Appendix~\ref{app:algorithm} gives the episode-level algorithm flow, and Appendix~\ref{app:training} gives the reward parsing and RL hyperparameters.

\subsection{Main Results}
\label{sec:main_results}

Table~\ref{tab:main_results} summarizes the results in all three benchmarks. Long-document performance is not determined by the backbone alone: GPT-5.4 rises from 62.4 direct accuracy on MMLongBench-Doc to 71.4 inside DocAtlas, a 9.0-point gain over the 65.8 human-expert reference. The gain appears across answerable evidence types, especially visual and structured cases, while unanswerable accuracy decreases from 75.6 to 68.6. This tradeoff is expected for an evidence-seeking agent: more retrieved evidence helps answerable questions but can make abstention harder when partially relevant pages are found. The same pattern appears beyond MMLongBench-Doc: GPT-5.4 improves from 55.1 to 75.6 LasJ on FinRAGBench-V and from 66.9 to 78.8 on LongDocURL; GPT-5.2 also reaches 70.6 on MMLongBench-Doc and 77.5 on LongDocURL.

In the fixed-reference auxiliary setting, the mutable environment also improves compact open policies. Qwen3.5-4B improves from 54.4 with direct input to 61.0 with DocAtlas, and Qwen3.5-9B improves from 58.0 to 61.6. The gains are strongest on tables and other structured evidence, where the agent can first localize candidate regions and then inspect a smaller set of pages or crops. This suggests that selective multimodal access helps compact models more than simply exposing the whole document in one context.

RL further improves compact policies beyond inference-time tool prompting. On MMLongBench-Doc, Qwen3.5-4B moves from 61.0 to 63.7 after RL, and Qwen3.5-9B from 61.6 to 64.4. The gains transfer beyond the RL data source: FinRAGBench-V improves from 67.9 to 71.7 for 4B and from 69.8 to 72.6 for 9B, while Qwen RL rows omit LongDocURL because it is used for data construction. The category changes indicate that RL is not only changing the final answer style; it improves the policy's choices about when to search, read, write notes, and review them. The GPT-5.4 auxiliary is not required by the method: \S\ref{sec:analysis} shows that open-weight or self auxiliaries keep most of the compact-policy gain.

\begin{figure*}[t]
\centering
\begin{minipage}[t]{0.48\textwidth}
\vspace{0pt}
\centering
\includegraphics[width=\linewidth]{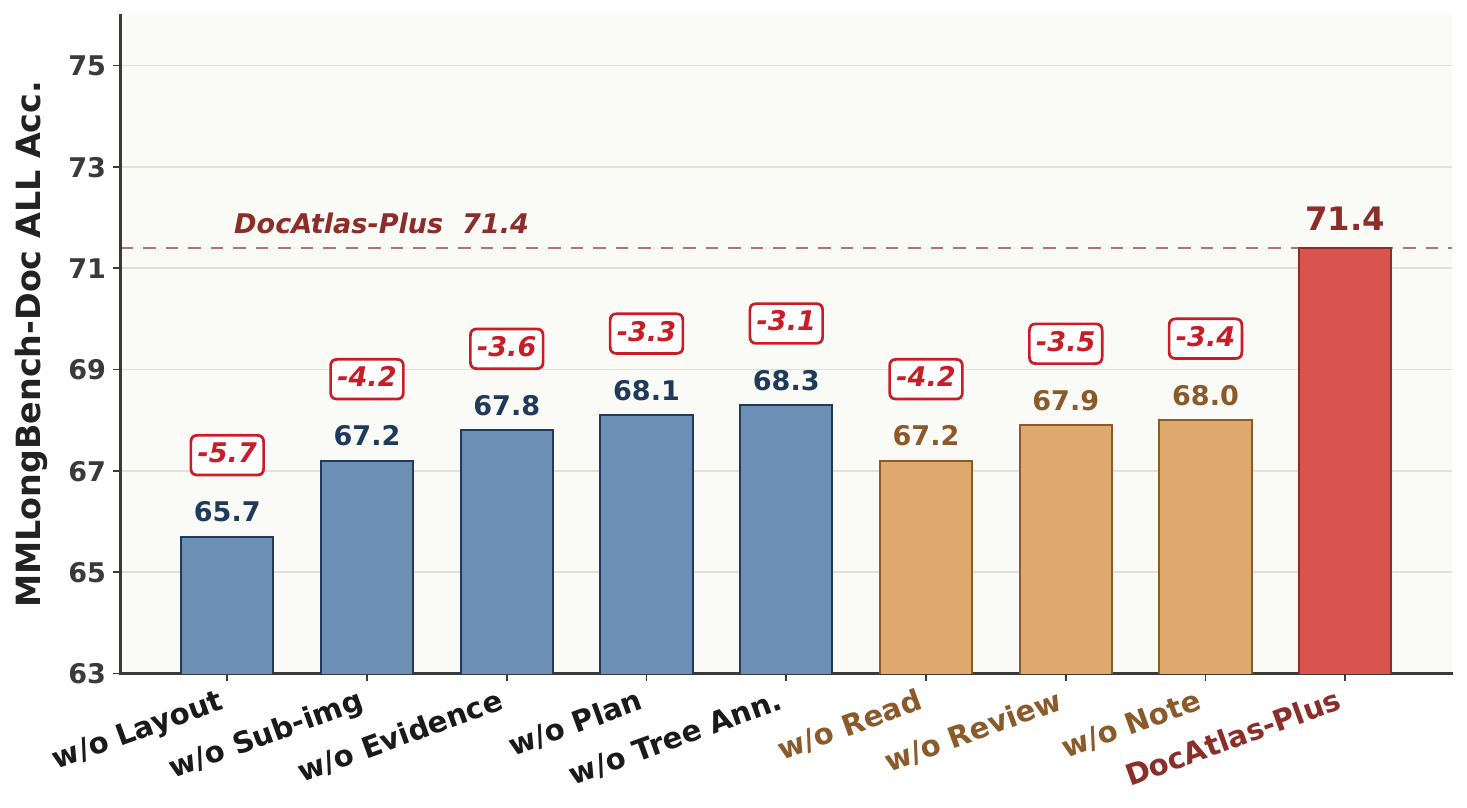}
\vspace{-3pt}
{\footnotesize (a) Component ablation.}
\end{minipage}
\hfill
\begin{minipage}[t]{0.48\textwidth}
\vspace{0pt}
\centering
\includegraphics[width=\linewidth]{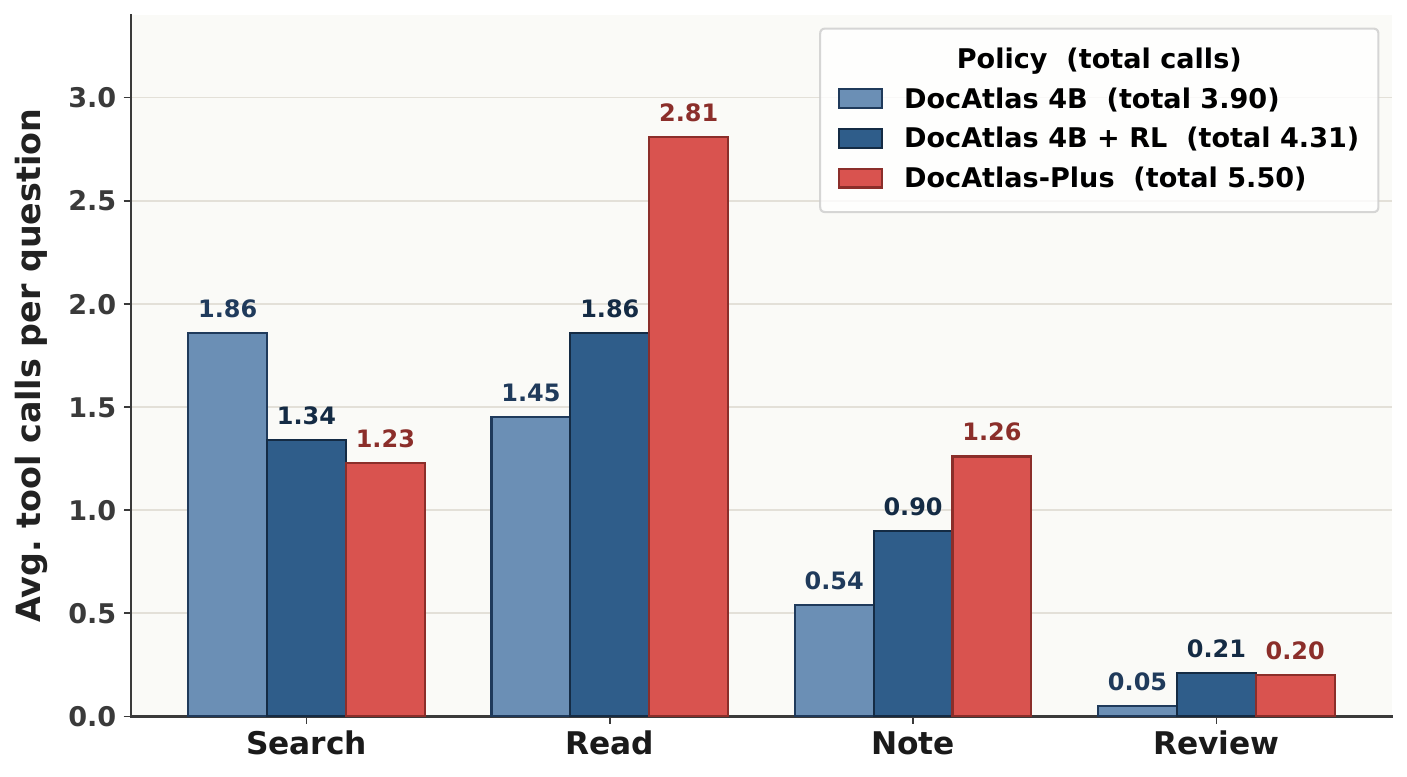}
\vspace{-3pt}
{\footnotesize (b) Average tool calls per question.}
\end{minipage}
\caption{Component and tool-use analysis on MMLongBench-Doc. (a) Bars report ALL accuracy after removing either an entire tool or an internal component of a tool; full DocAtlas with GPT-5.4 reaches 71.4. Detailed category-level results are in Appendix~\ref{app:ablation}. (b) After RL, the 4B policy uses fewer \textsc{Search} calls and more \textsc{Read}, \textsc{Note}, and \textsc{Review} calls despite only a small increase in total calls; DocAtlas-Plus denotes the GPT-5.4 instantiation.}
\label{fig:ablation_analysis}
\label{fig:tool_usage}
\end{figure*}
\subsection{Ablation Study}
\label{sec:ablation}

Figure~\ref{fig:ablation_analysis}(a) ablates DocAtlas on MMLongBench-Doc; Appendix~\ref{app:ablation} gives the full category-level table. The largest losses come from visual access. Removing layout images lowers ALL from 71.4 to 65.7, with large drops on figures and charts; removing cropped sub-images lowers ALL to 67.2. Full-page layout images preserve spatial context and cross-element relations, while cropped sub-images expose local details inside figures, charts, and tables. The result supports the two-level design of \textsc{Read}: neither text-only extraction nor a single visual scale is sufficient for long documents with complex layouts and visual content.

Selective evidence access also matters. Without \textsc{Read}, the agent consumes \textsc{Search} candidates directly and ALL falls to 67.2, showing that \textsc{Search} should remain a high-recall proposal mechanism rather than the final evidence selector. The mutable-state components each contribute consistent gains: removing the evidence field, \textsc{Review}, \textsc{Note}, plan field, or tree annotation yields 67.8--68.3 ALL. These similar drops show that memory is a pathway rather than one switch: notes preserve page-grounded facts, review retrieves them across steps, and tree annotations feed local findings back into later search.
\subsection{Behavior and Robustness Analysis}
\label{sec:analysis}
\begin{wraptable}{r}{0.52\linewidth}
\vspace{-8pt}
\centering
\caption{Search-Read evidence efficiency on MMLongBench-Doc. All-Hit is the fraction of examples for which all gold evidence pages are covered; F1 is page-level evidence F1.}
\small
\setlength\tabcolsep{5.2pt}
\renewcommand{\arraystretch}{1.06}
\begin{tabular}{lccc}
\toprule
\textbf{Method} & \textbf{Avg.} & \textbf{All-Hit} & \textbf{F1} \\
\midrule
ColQwen top-2  & 2.00  & 64.12 & 38.75 \\
ColQwen top-6  & 6.00  & 76.42 & 24.36 \\
ColQwen top-10 & 10.00 & 83.60 & 18.38 \\
\midrule
Full \textsc{Search} & 16.65 & \textbf{87.45} & 38.33 \\
\quad w/o \textsc{Annotation} & 16.94 & 80.05 & 29.40 \\
First \textsc{Read} & 2.67 & 59.90 & 56.04 \\
\quad w/o \textsc{Annotation} & 2.61 & 58.21 & 56.37 \\
Full \textsc{Read} & 5.79 & 78.01 & \textbf{58.99} \\
\quad w/o \textsc{Annotation} & 6.01 & 76.93 & 58.17 \\
\bottomrule
\end{tabular}
\label{tab:retrieval_efficiency}
\vspace{-10pt}
\end{wraptable}
This section checks whether the gains in Table~\ref{tab:main_results} come from the intended behavior: learned tool use, efficient evidence access, and robustness to the auxiliary model.

\textbf{Tool-use behavior.}
Figure~\ref{fig:tool_usage} (b) shows that RL changes the compact policy's behavior without simply increasing its tool budget. Total calls grow by only 10.5\%, but the mix shifts sharply: \textsc{Search} decreases by 28.0\%, while \textsc{Read}, \textsc{Note}, and \textsc{Review} increase by 28.3\%, 66.7\%, and 320.0\%. The policy is not just trying more actions; it shifts effort from broad localization to reading pages, recording evidence, and checking saved notes. The clearest change is \textsc{Review}: the no-RL 4B policy almost never uses it (0.05 calls/question), whereas the RL policy reaches 0.21, matching GPT-5.4's 0.20 rate. This suggests that RL teaches the compact model to use the harness's note-and-verify pathway rather than treating each read as an isolated context extension.

\textbf{Evidence efficiency.}
Table~\ref{tab:retrieval_efficiency} explains why \textsc{Search} and \textsc{Read} are separate tools. Raw tree search has high coverage (87.45 All-Hit) but is intentionally broad, returning 16.65 pages on average. This is useful for recall but too expensive to pass directly to the policy. The full reading trajectory inspects far fewer pages (5.79) while achieving the best page-level F1 (58.99), showing that \textsc{Read} converts high-recall candidates into a compact evidence set. ColQwen shows the opposite tradeoff: larger top-$k$ raises All-Hit but lowers F1 by retrieving many extra pages. Tree annotations also matter: removing them drops raw-search All-Hit from 87.45 to 80.05 and page F1 from 38.33 to 29.40, so the mutable tree improves later search rather than only storing notes for final synthesis.

\begin{wraptable}{r}{0.48\linewidth}
\vspace{-8pt}
\centering
\caption{Auxiliary LLM sensitivity on MMLongBench-Doc. The RL policy is fixed; only the frozen model used by \textsc{Search}/\textsc{Review} is varied. ``Self'' uses the corresponding Qwen policy model as the auxiliary. $\Delta$ is relative to GPT-5.4.}
\label{tab:aux_sensitivity}
\small
\setlength\tabcolsep{5.2pt}
\renewcommand{\arraystretch}{1.06}
\begin{tabular}{lcccc}
\toprule
\textbf{Aux.} & \textbf{4B} & \textbf{$\Delta$} & \textbf{9B} & \textbf{$\Delta$} \\
\midrule
GPT-5.4 & 63.7 & 0.0 & 64.4 & 0.0 \\
Qwen3.5-35B-A3B & 63.4 & -0.3 & 64.0 & -0.4 \\
Self & 62.0 & -1.7 & 63.2 & -1.2 \\
\bottomrule
\end{tabular}
\vspace{-10pt}
\end{wraptable}
\textbf{Auxiliary robustness.}
Finally, we test whether the compact-policy gains simply come from using GPT-5.4 inside \textsc{Search} and \textsc{Review}. Holding the RL policy fixed, Table~\ref{tab:aux_sensitivity} replaces only this auxiliary model. The GPT-5.4 row is a reference setting, not a required deployment choice: Qwen3.5-35B-A3B loses only 0.3--0.4 points, and ``Self'' keeps both policies within 2 points. The auxiliary routes search and selects saved notes, but the trained policy still decides when to invoke tools, which pages to read, and how to answer. The fully open-weight self setting still reaches 62.0/63.2 for the 4B/9B policies, suggesting that DocAtlas is not just a wrapper around a proprietary auxiliary model.

\section{Conclusion}
\label{sec:conclusion}
DocAtlas treats long-document understanding as mutable-state interaction: agents search a visual-aware tree, read pages, write grounded notes, and review prior findings. This turns a long document into state that the policy updates while it works. Across three benchmarks, DocAtlas improves large VLM agents and helps compact Qwen policies learn tool use through RL. The gains come from multimodal reading, decoupled \textsc{Search}/\textsc{Read}, mutable notes, tree annotations, and \textsc{Search}/\textsc{Review} auxiliaries that can be replaced by open-weight or self models. Overall, long-document understanding depends not only on context length, but also on how the interface lets a model find, remember, and use evidence.

\bibliographystyle{plain}
\bibliography{references}
  
\newpage
\appendix

\section{Additional Experimental Results}

\subsection{Full Ablation Results}
\label{app:ablation}
Table~\ref{tab:ablation_mmlongbench} reports the category-level ablation results corresponding to Figure~\ref{fig:ablation_analysis}.

\begin{table}[h]
\centering
\caption{Ablation study of DocAtlas with GPT-5.4 as the agent, \textsc{Search}, and \textsc{Review} LLM on MMLongBench-Doc. Each row removes one component from the full system. Evidence-source columns are text (TXT), layout (LAY), chart (CHA), table (TAB), figure (FIG), and unanswerable (UNA); Acc is overall accuracy and F1 is overall F1. }
\small
\renewcommand{\arraystretch}{1.08}
\setlength\tabcolsep{8pt}
\begin{tabular}{@{}l cccccc cc@{}}
\toprule
\textbf{Variant} & \textbf{TXT} & \textbf{LAY} & \textbf{CHA} & \textbf{TAB} & \textbf{FIG} & \textbf{UNA} & \textbf{Acc} & \textbf{F1} \\
\midrule
w/o \textsc{Read}         & 63.9 & 67.4 & 63.5 & 71.9 & 62.8 & 69.1 & 67.2 & 66.1 \\
w/o \textsc{Note}         & 65.8 & 72.5 & 65.3 & 71.8 & 63.4 & 67.9 & 68.0 & 66.9 \\
w/o \textsc{Review}       & 64.9 & 71.7 & 62.9 & 71.3 & 62.5 & 71.8 & 67.9 & 66.4 \\
w/o layout image          & 64.1 & 62.3 & 63.1 & 71.6 & 58.8 & 67.4 & 65.7 & 64.9 \\
w/o sub-image             & 62.0 & 70.0 & 62.9 & 73.4 & 63.3 & 68.6 & 67.2 & 65.8 \\
w/o evidence field        & 64.8 & 71.6 & 62.4 & 70.1 & 64.2 & 69.4 & 67.8 & 66.8 \\
w/o plan field            & 65.9 & 66.7 & 63.8 & 70.8 & 65.6 & 68.8 & 68.1 & 67.1 \\
w/o tree annotation       & 64.7 & 71.6 & 64.0 & 72.5 & 63.9 & 68.6 & 68.3 & 67.4 \\
\midrule
\textbf{Full DocAtlas} & \textbf{68.8} & \textbf{69.0} & \textbf{68.2} & \textbf{74.1} & \textbf{69.8} & \textbf{68.6} & \textbf{71.4} & \textbf{70.3} \\
\bottomrule
\end{tabular}
\label{tab:ablation_mmlongbench}
\end{table}

\subsection{Full Tool-Usage Statistics}
\label{app:tool_usage}
Table~\ref{tab:tool_usage_full} reports the raw tool-call statistics underlying Figure~\ref{fig:tool_usage}. 

\begin{table}[h]
\centering
\caption{Average tool calls per question on MMLongBench-Doc. Total is the sum of the four normalized tool aliases: \textsc{Search}, \textsc{Read}, \textsc{Note}, and \textsc{Review}.}
\small
\setlength\tabcolsep{5pt}
\renewcommand{\arraystretch}{1.08}
\begin{tabular}{lrrrrr}
\toprule
\textbf{Policy} & \textbf{Search} & \textbf{Read} & \textbf{Note} & \textbf{Review} & \textbf{Total} \\
\midrule
DocAtlas-Plus (GPT-5.4)  & 1.23 & 2.81 & 1.26 & 0.20 & 5.50 \\
DocAtlas-4B (Qwen3.5-4B)  & 1.86 & 1.45 & 0.54 & 0.05 & 3.90 \\
DocAtlas-4B-RL   & 1.34 & 1.86 & 0.90 & 0.21 & 4.31 \\
\bottomrule
\end{tabular}
\label{tab:tool_usage_full}
\end{table}

Table~\ref{tab:tool_usage_shift} summarizes the relative change induced by RL for the 4B policy. RL increases the total tool budget by only 10.5\%, but clearly changes the tool mix: \textsc{Search} decreases while \textsc{Read}, \textsc{Note}, and \textsc{Review} increase.

\begin{table}[h]
\centering
\caption{Tool-use shift from DocAtlas-4B to DocAtlas-4B-RL on MMLongBench-Doc.}
\small
\setlength\tabcolsep{8pt}
\renewcommand{\arraystretch}{1.08}
\begin{tabular}{lrrrr}
\toprule
\textbf{Tool} & \textbf{4B no-RL} & \textbf{4B-RL} & \textbf{Abs. change} & \textbf{Rel. change} \\
\midrule
\textsc{Search} & 1.86 & 1.34 & -0.52 & -28.0\% \\
\textsc{Read}   & 1.45 & 1.86 & +0.41 & +28.3\% \\
\textsc{Note}   & 0.54 & 0.90 & +0.36 & +66.7\% \\
\textsc{Review} & 0.05 & 0.21 & +0.16 & +320.0\% \\
\midrule
Total & 3.90 & 4.31 & +0.41 & +10.5\% \\
\bottomrule
\end{tabular}
\label{tab:tool_usage_shift}
\end{table}

\clearpage
\subsection{Algorithm Flow}
\label{app:algorithm}

The pseudocode below gives a compact view of one DocAtlas episode. The same loop is used for inference and for RL rollouts; RL additionally stores the completed trajectory and scores it with the reward in Appendix~\ref{app:training}. The pseudocode abstracts away implementation-specific tool schemas, but matches the state variables and tools defined in Section~\ref{sec:method}.

\begin{algorithm}[h]
\caption{DocAtlas mutable-state interaction loop.}
\label{alg:docatlas_episode}
\small
\begin{algorithmic}[1]
\Require document $D=\{d_1,\ldots,d_N\}$, question $q$, initial tree $G^0$, frozen tools \textsc{Search}, \textsc{Read}, \textsc{Note}, \textsc{Review}, policy $\pi_\theta$, maximum turns $T_{\max}$
\State Initialize note store $M^0 \gets \emptyset$, explored pages $H^0 \gets \emptyset$, and history $h_1 \gets [q, \mathrm{overview}(G^0)]$
\For{$t=1$ \textbf{to} $T_{\max}$}
    \State $u_t \gets \pi_\theta(h_t)$
    \If{$u_t = \textsc{Finish}(a)$}
        \State \Return answer $a$ and trajectory $\tau$
    \ElsIf{$u_t = \textsc{Search}(z)$}
        \State $o_t \gets \textsc{Search}(z; G^t, H^t)$
        \State $(G^{t+1},M^{t+1},H^{t+1}) \gets (G^t,M^t,H^t)$
    \ElsIf{$u_t = \textsc{Read}(P)$}
        \State $o_t \gets \textsc{Read}(P)$
        \State $(G^{t+1},M^{t+1},H^{t+1}) \gets (G^t,M^t,H^t \cup P)$
    \ElsIf{$u_t = \textsc{Note}(b,e,p)$}
        \State $M^{t+1} \gets M^t \cup \{(b,e,p)\}$; \quad $G^{t+1} \gets \mathrm{Enrich}(G^t,e)$; \quad $H^{t+1} \gets H^t$
        \State $o_t \gets$ acknowledgement
        \If{archival is enabled}
            \State Replace bulky prior \textsc{Read} observations with short placeholders
        \EndIf
    \ElsIf{$u_t = \textsc{Review}(z)$}
        \State $C_t \gets \mathrm{Cards}(M^t)$; \quad $o_t \gets \textsc{Review}(z,C_t)$
        \State $(G^{t+1},M^{t+1},H^{t+1}) \gets (G^t,M^t,H^t)$
    \Else
        \State $o_t \gets$ invalid-tool message
        \State $(G^{t+1},M^{t+1},H^{t+1}) \gets (G^t,M^t,H^t)$
    \EndIf
    \State $h_{t+1} \gets h_t \mathbin{\Vert} (u_t,o_t)$
\EndFor
\State \Return timeout answer and trajectory $\tau$
\Statex
\Statex \textbf{RL scoring.} For a completed trajectory, set $r=0$ if the final answer is not boxed; otherwise extract the boxed content and set
\Statex \hspace{1.5em}$r = \mathrm{Score}_{\mathrm{LongDocURL}}\bigl(\mathrm{ExtractBoxed}(a),a^\star\bigr)$.
\end{algorithmic}
\end{algorithm}

\subsection{Training and Reward Details}
\label{app:training}

Table~\ref{tab:training_config} summarizes the RL configuration used for the compact Qwen3.5 policies. We use the same DocAtlas tool environment as in inference, keep the tree, tool executors, and auxiliary \textsc{Search}/\textsc{Review} models frozen, and train only the VLM policy. The implementation uses verl with vLLM multi-turn rollout and FSDP2 actor/reference workers. Cloud-storage paths, internal cluster identifiers, and credentials are not part of the method and are omitted.

\begin{table}[h]
\centering
\caption{RL training configuration for DocAtlas compact-policy training. Values reflect the default DaPO run used for the Qwen3.5-4B results unless stated otherwise.}
\small
\setlength\tabcolsep{4pt}
\renewcommand{\arraystretch}{1.08}
\begin{tabular}{@{}p{0.27\linewidth}p{0.68\linewidth}@{}}
\toprule
\textbf{Category} & \textbf{Setting} \\
\midrule
Base policy & Qwen3.5 VLM initialized from the local checkpoint \\
Algorithm & DAPO-style GRPO; group-normalized advantages \\
Rollouts & $n{=}8$ sampled trajectories per prompt \\
Epochs and batch size & 1 epoch; train batch size 8; 8 GPUs per node \\
Learning rates & actor learning rate $1\times10^{-6}$; critic learning rate $1\times10^{-6}$ \\
KL and entropy & token-level KL coefficient 0.01; low-variance KL; entropy coefficient 0 \\
Clipping & asymmetric DAPO clipping with $\epsilon_{\mathrm{low}}{=}0.1$, $\epsilon_{\mathrm{high}}{=}0.3$ \\
Sequence limits & max prompt length 8192; max response length 16384 \\
Rollout engine & vLLM; tensor parallel size 1; GPU memory utilization 0.85 \\
Software stack & Python 3.12; PyTorch 2.10; vLLM 0.18; Transformers $\geq$5.5; Ray $\geq$2.54; flash-attn 2.8 \\
\bottomrule
\end{tabular}
\label{tab:training_config}
\end{table}

\textbf{Reward function.}
For RL training, the reward is computed from the final generated answer. The last visible answer is the text after the last \texttt{</think>} marker, if present. If this answer does not contain a parsable \texttt{\textbackslash boxed\{\}} field, the reward is 0. Otherwise, we extract the content inside the final boxed answer and apply the official LongDocURL type-aware evaluation score:
\begin{equation}
    r = \mathrm{Score}_{\mathrm{LongDocURL}}\bigl(\mathrm{ExtractBoxed}(a), a^\star\bigr).
\end{equation}
The score is computed only from the extracted boxed answer and the ground-truth answer; no separate process term is used.

\subsection{Qualitative Trajectory Examples}
\label{app:case_studies}
\begin{figure*}[t]
\centering
\includegraphics[width=\linewidth]{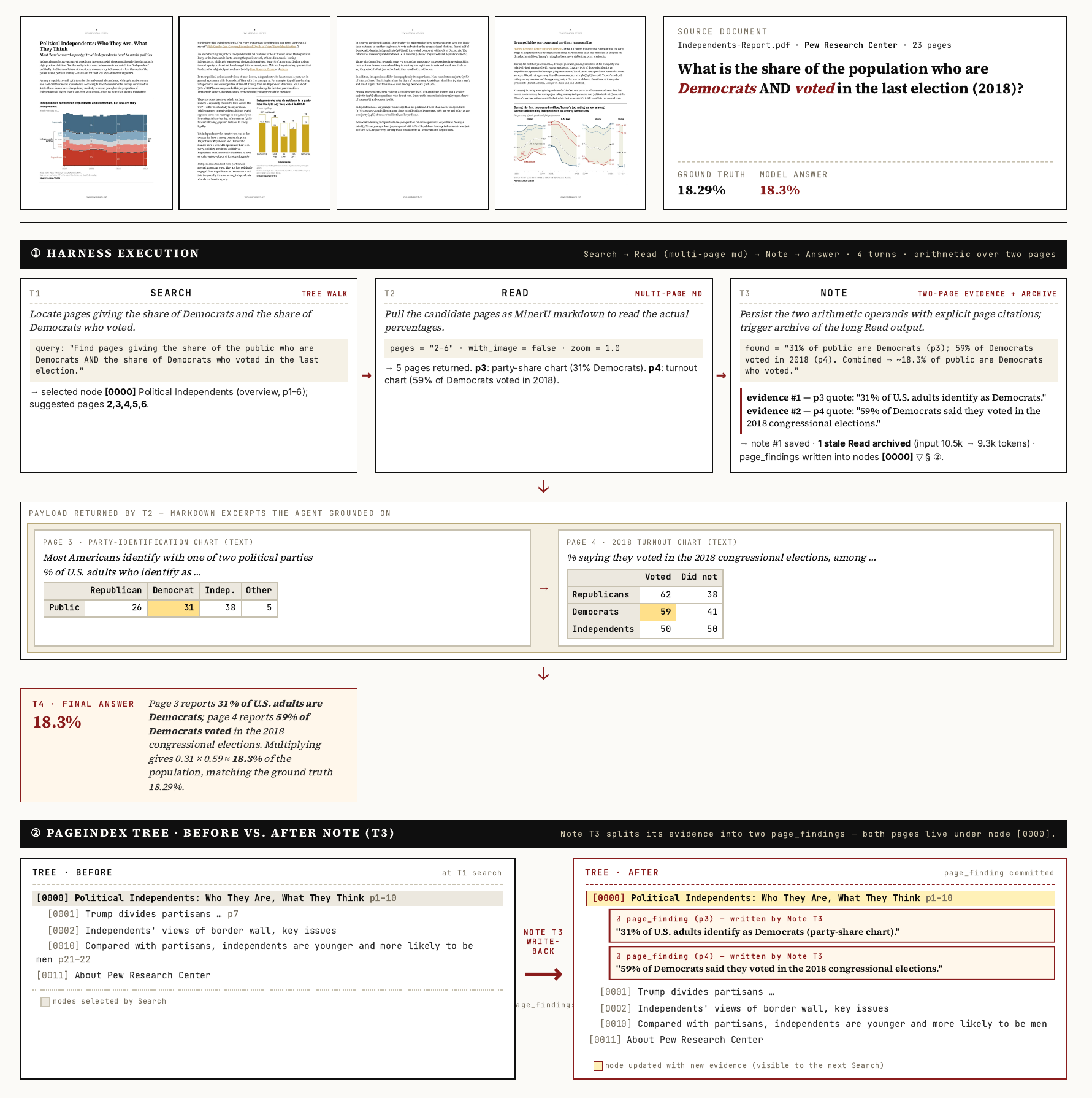}
\caption{Qualitative trajectory example: multi-hop evidence gathering. DocAtlas iteratively searches the document tree, reads selected evidence pages, records intermediate findings in notes, and combines them through review before answering.}
\label{fig:case_multihop}
\end{figure*}

\begin{figure*}[t]
\centering
\includegraphics[width=\linewidth]{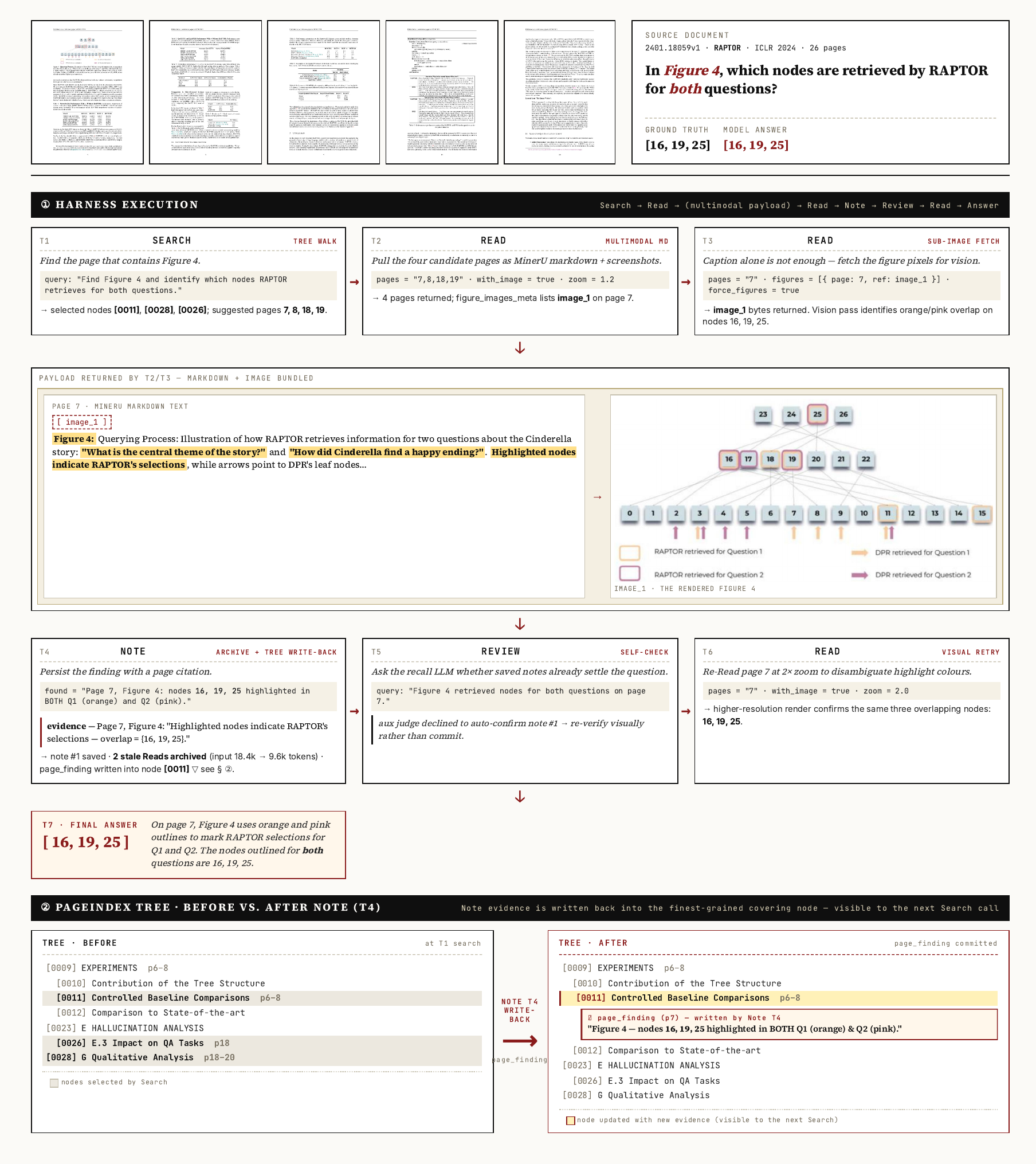}
\caption{Qualitative trajectory example: complete DocAtlas reasoning flow. The trajectory shows how tree search, selective multimodal reading, source-grounded notes, and final evidence consolidation interact within one episode.}
\label{fig:case_raptor}
\end{figure*}

Figure~\ref{fig:case_multihop} and Figure~\ref{fig:case_raptor} show two complete DocAtlas reasoning trajectories. These examples illustrate the full interaction loop: the policy first uses \textsc{Search} to localize a broad candidate region in the mutable document tree, then calls \textsc{Read} on selected pages or visual regions, writes source-grounded \textsc{Note} entries, optionally uses \textsc{Review} to recall earlier findings, and finally produces a page-grounded answer. The examples are intended to make the environment dynamics concrete rather than serve as additional quantitative evidence.

\input{interaction_prompts}

\section{Broader Impacts, Limitations, and LLM Usage}
\label{app:impact_limitations_llm}

\textbf{Broader impacts.}
DocAtlas is intended to improve access to information in long documents with complex layouts and visual content, such as reports, scientific articles, manuals, filings, and policy documents. Potential positive impacts include reducing the cost of searching and cross-checking evidence in large document collections, improving accessibility for users who need to inspect complex documents, and enabling smaller open-weight VLMs to perform more useful document analysis through tool-mediated interaction instead of relying only on larger backbones. At the same time, the system may be misused to summarize or extract information from sensitive or copyrighted documents without appropriate authorization. It may also produce overconfident answers when retrieved evidence is incomplete or when the source document itself is ambiguous, outdated, or misleading. For high-stakes uses such as legal, medical, financial, or policy decisions, DocAtlas should be treated as an evidence-navigation aid rather than an autonomous decision maker; users should inspect the cited pages and apply domain-specific review.

\textbf{Limitations.}
Our evaluation focuses on existing long-document understanding benchmarks and may not cover all document genres, languages, scan qualities, or domain-specific reasoning patterns encountered in deployment. DocAtlas relies on document preprocessing, including markdown extraction, visual parsing, page-image rendering, cropped sub-images, and a question-agnostic tree. Errors in these preprocessing steps can affect downstream search and reading. The environment also uses auxiliary LLMs for \textsc{Search} and \textsc{Review}. We use GPT-5.4 as a fixed reference auxiliary in the main comparison, and the sensitivity study shows that open-weight and self auxiliaries remain close; still, performance and cost can depend on the quality of these frozen components. The current RL setting trains compact Qwen policies with sparse outcome rewards and pass@16-filtered data from LongDocURL, so further work is needed to understand scaling behavior, reward design, training stability, and transfer to broader domains. Finally, DocAtlas improves evidence access and memory, but it does not guarantee factual correctness: the agent can still miss relevant pages, misread visual elements, write incomplete notes, or combine evidence incorrectly.

\textbf{Declaration of LLM usage.}
Large language models are part of the research object studied in this paper: DocAtlas uses VLM agents, auxiliary LLMs for \textsc{Search} and \textsc{Review}, and LLM-as-judge evaluation following the benchmark protocols described in the main text. In addition, the authors used LLM-based writing assistance only to check grammar and polish wording of the manuscript. This writing assistance did not generate new experimental results, change the methodology, select data, perform analysis, or alter the scientific claims; all technical content, experiments, results, and conclusions were reviewed and approved by the authors.

\newpage

\end{document}

%% file: interaction_prompts.tex
\subsection{DocAtlas Prompt Design}
\label{app:prompts}

This appendix summarizes the prompt material and tool-call contracts used by DocAtlas. The runtime system prompt is assembled from a fixed preamble, tool-use protocol, optional memory and tree-annotation blocks, the faithfulness rule, the final-answer format, and the four skill descriptions. 

\subsubsection{System prompt template}
\label{app:prompts:system}

\begin{tcolorbox}[colback=blue!4,colframe=blue!35,title=\textbf{Document context and role},breakable]
\small
You are an expert document analysis assistant. Answer user questions by strictly following the tool usage and reasoning protocol. The document has been pre-indexed into a hierarchical tree structure similar to a detailed table of contents. The user message provides a lightweight table of contents with node IDs, titles, and page ranges, but no authoritative section content. The agent must use tools to inspect actual document evidence.
\end{tcolorbox}

\begin{tcolorbox}[colback=blue!4,colframe=blue!35,title=\textbf{Tool strategy},breakable]
\small
\textbf{Search} is the coarse filter: it performs structural tree search over the document hierarchy to locate relevant sections or page ranges. Queries should be specific natural-language descriptions rather than short keywords.\\[2pt]
\textbf{Read} is the fine filter: it reads selected pages and returns markdown text, layout images, and optional cropped sub-images. The agent should read only needed pages, normally at most five pages per call.\\[2pt]
\textbf{Note} records established findings, source-grounded evidence, and the remaining plan. It acts as a compact reasoning checkpoint.\\[2pt]
\textbf{Review} recalls previously saved notes using a focused query. It returns only matched notes rather than a full transcript dump.
\end{tcolorbox}

\begin{tcolorbox}[colback=blue!4,colframe=blue!35,title=\textbf{Mandatory tool-use protocol},breakable]
\small
The agent must call at least one tool before answering. The tree overview alone is insufficient; factual answers must be verified by reading page content. For specific questions, the default flow is \textsc{Search} $\rightarrow$ \textsc{Read} $\rightarrow$ answer. For broad questions, the agent should search for key sections and read representative pages. When search returns multiple candidates, the agent may start from the most promising subset, but should not skip unread candidates unless it can explain why the candidate set is structurally insufficient. Page numbers in questions are treated cautiously because printed page or slide numbers may differ from physical PDF pages; search is used to resolve this mapping before reading.
\end{tcolorbox}

\begin{tcolorbox}[colback=green!4,colframe=green!35!black,title=\textbf{Note and Review protocol},breakable]
\small
The agent writes a \textsc{Note} whenever it finds useful information: confirmed facts, key numbers, page references, negative findings, or changes in search strategy. Every note should contain source-grounded evidence rather than only a high-level summary. For multi-hop or cross-page questions, notes preserve partial findings so later reasoning does not rely on long transcript context.\\[2pt]
The agent calls \textsc{Review} when the answer requires combining evidence from multiple notes, when it needs to recall earlier findings, or before answering \emph{Not answerable} after a multi-step search. Review is skipped for simple single-page answers where all evidence is contained in the latest read result.
\end{tcolorbox}

\begin{tcolorbox}[colback=green!4,colframe=green!35!black,title=\textbf{Memory management and tree annotation},breakable]
\small
When memory management is enabled, calling \textsc{Note} archives earlier bulky \textsc{Read} observations into compact placeholders while preserving source-grounded evidence in the note store. Archived pages can be re-read if needed.\\[2pt]
When tree annotation is enabled, notes with explicit page references write page-level findings back into the finest-grained tree nodes covering those pages. These \texttt{page\_findings} are partial observations from prior query-driven reads, not complete page summaries. Later \textsc{Search} calls can use them as auxiliary hints while still relying primarily on the original tree structure.
\end{tcolorbox}

\begin{tcolorbox}[colback=red!3,colframe=red!35,title=\textbf{Faithfulness and abstention rule},breakable]
\small
The agent must be faithful to the provided pages. If the read evidence does not contain sufficient information, the final answer must be exactly \texttt{Not answerable}. When abstaining, the final-answer line must contain only \texttt{Not answerable}; the reasoning may explain why the document lacks the evidence, but must not offer a guess or answer a related question.
\end{tcolorbox}

\begin{tcolorbox}[colback=gray!6,colframe=black!30,title=\textbf{Final-answer format},breakable]
\small
The final response must use the benchmark format:\\[2pt]
\texttt{Final answer: <shortest exact span>}\\
\texttt{Reasoning: <1--3 sentence explanation with page references>}\\[2pt]
The final-answer line should contain only the answer. Numeric answers drop unit words unless required. List answers use a one-line Python list literal and preserve the document wording. Unanswerable cases use exactly \texttt{Final answer: Not answerable}.
\end{tcolorbox}

\subsubsection{Per-skill prompt material}
\label{app:prompts:skills}

\begin{tcolorbox}[colback=orange!5,colframe=orange!45!black,title=\textbf{Search skill},breakable]
\small
\textsc{Search} is the discovery step. Given a question, an auxiliary LLM walks the document tree and selects nodes likely to contain the answer. The harness expands selected nodes into suggested page ranges and records search history to avoid repeating the same pages. Search does not read content; it only proposes where to inspect next. Good queries name the target entity, aspect, and constraints, e.g., ``Find sections discussing partisan splits on presidential ethics and transparency,'' rather than a keyword such as ``ethics.''
\end{tcolorbox}

\begin{tcolorbox}[colback=orange!5,colframe=orange!45!black,title=\textbf{Read skill},breakable]
\small
\textsc{Read} is the only tool that brings document content into the conversation. It fetches selected PDF pages as MinerU markdown or PyPDF text, with optional full-page screenshots and cropped sub-images. It is used after \textsc{Search} identifies candidate pages, when the actual page text or visual content is needed. The prompt discourages reading more than five pages per call and encourages selective follow-up reads for charts, figures, tables, or scanned pages.
\end{tcolorbox}

\begin{tcolorbox}[colback=orange!5,colframe=orange!45!black,title=\textbf{Note skill},breakable]
\small
\textsc{Note} appends a progress-analysis note containing what was found, what remains to do, and evidence linked to page references. The prompt requires exact quoted evidence when possible and discourages unsupported summaries. Notes let the agent preserve important findings without carrying all page observations in active context; later \textsc{Review} calls can retrieve relevant note bodies by query.
\end{tcolorbox}

\begin{tcolorbox}[colback=orange!5,colframe=orange!45!black,title=\textbf{Review skill},breakable]
\small
\textsc{Review} recalls saved notes by query. The harness builds compact note cards, an auxiliary LLM selects relevant cards, and the tool returns the full selected note bodies. Review is used before final answers that combine multiple findings, before a new search when earlier notes may already contain the answer, or before abstaining after a multi-step search. It is not used when no notes exist or when the answer depends only on the latest single read result.
\end{tcolorbox}

\subsubsection{Tool-call schemas}
\label{app:prompts:schemas}

\begin{tcolorbox}[colback=purple!4,colframe=purple!35,title=\textbf{Search schema},breakable]
\small
\textbf{Purpose.} Locate relevant document-tree nodes and return suggested physical pages.\\[2pt]
\begin{tabular}{@{}lll@{}}
\toprule
\textbf{Field} & \textbf{Type} & \textbf{Description} \\
\midrule
\texttt{query} & string, required & Detailed natural-language search query. \\
\bottomrule
\end{tabular}
\vspace{3pt}

\textit{Contract.} The query should specify the entity, aspect, and constraints; short keyword queries are discouraged because tree search is LLM-guided rather than lexical matching.
\end{tcolorbox}

\begin{tcolorbox}[colback=purple!4,colframe=purple!35,title=\textbf{Read schema},breakable]
\small
\textbf{Purpose.} Fetch page text, page screenshots, or selected sub-images.\\[2pt]
\begin{tabular}{@{}lll@{}}
\toprule
\textbf{Field} & \textbf{Type} & \textbf{Description} \\
\midrule
\texttt{pages} & string, required & Pages or ranges, e.g., \texttt{1,3-5,8}. \\
\texttt{with\_image} & boolean & Attach full-page screenshots. \\
\texttt{figures} & list & Fetch sub-images by \texttt{(page, ref)} from prior metadata. \\
\texttt{force\_figures} & boolean & Bypass the minimum-size filter for requested figures. \\
\texttt{zoom} & number & Zoom factor for page screenshots. \\
\texttt{doc\_id} & string & Optional document identifier. \\
\bottomrule
\end{tabular}
\vspace{3pt}

\textit{Contract.} \texttt{pages} is the only required field. The prompt recommends no more than five pages per call and uses \texttt{figures} only after a previous read exposes a \texttt{figure\_images\_meta} catalog.
\end{tcolorbox}

\begin{tcolorbox}[colback=purple!4,colframe=purple!35,title=\textbf{Note schema},breakable]
\small
\textbf{Purpose.} Save a compact, source-grounded reasoning checkpoint and optionally trigger memory/tree side effects.\\[2pt]
\begin{tabular}{@{}lll@{}}
\toprule
\textbf{Field} & \textbf{Type} & \textbf{Description} \\
\midrule
\texttt{found} & string, required & Short summary of established findings. \\
\texttt{plan} & string & Remaining gaps and next intended step. \\
\texttt{evidence} & list & Source-grounded evidence entries. \\
\texttt{side\_effect\_policy} & enum & Optional archive/enrich behavior override. \\
\bottomrule
\end{tabular}
\vspace{3pt}

\textbf{Evidence entry.}
\begin{tabular}{@{}lll@{}}
\toprule
\textbf{Field} & \textbf{Type} & \textbf{Description} \\
\midrule
\texttt{type} & enum & \texttt{text}, \texttt{table}, or \texttt{image}. \\
\texttt{source} & string, required & Explicit page reference, e.g., \texttt{Page 5}. \\
\texttt{content} & string & Quoted text, table row, or caption. \\
\texttt{filename} & string & Image filename for image evidence. \\
\bottomrule
\end{tabular}
\vspace{3pt}

\textit{Contract.} Notes should contain page-grounded evidence. When tree annotation is enabled, page references in evidence allow the harness to write findings back into the document tree.
\end{tcolorbox}

\begin{tcolorbox}[colback=purple!4,colframe=purple!35,title=\textbf{Review schema},breakable]
\small
\textbf{Purpose.} Recall previously saved notes by query.\\[2pt]
\begin{tabular}{@{}lll@{}}
\toprule
\textbf{Field} & \textbf{Type} & \textbf{Description} \\
\midrule
\texttt{query} & string, required & Focused description of what to recall from notes. \\
\bottomrule
\end{tabular}
\vspace{3pt}

\textit{Contract.} Review searches only saved notes, not document pages. It returns matched note bodies selected by an auxiliary LLM.
\end{tcolorbox}

%% file: references.bib
@article{cho2024m3docrag,
  title={M3docrag: Multi-modal retrieval is what you need for multi-page multi-document understanding},
  author={Cho, Jaemin and Mahata, Debanjan and Irsoy, Ozan and He, Yujie and Bansal, Mohit},
  journal={arXiv preprint arXiv:2411.04952},
  year={2024}
}

@article{gong2025mhier,
  title={MHier-RAG: Multi-Modal RAG for Visual-Rich Document Question-Answering via Hierarchical and Multi-Granularity Reasoning},
  author={Gong, Ziyu and Mai, Chengcheng and Huang, Yihua},
  journal={arXiv preprint arXiv:2508.00579},
  year={2025}
}

@article{gomes2025visdocsketcher,
  title={VisDocSketcher: Towards Scalable Visual Documentation with Agentic Systems},
  author={Gomes, Lu{\'\i}s F and Zhou, Xin and Lo, David and Abreu, Rui},
  journal={arXiv preprint arXiv:2509.11942},
  year={2025}
}

@inproceedings{chen2025hear,
  title={HEAR: A Holistic Extraction and Agentic Reasoning Framework for Document Understanding},
  author={Chen, Longfeng and Xiao, Zheng and Wang, Juyuan and Huang, Zeyu and Zeng, Yawen and Xu, Jin},
  booktitle={ACM MM},
  pages={14376--14382},
  year={2025}
}

@article{li2026deepread,
  title={DeepRead: Document Structure-Aware Reasoning to Enhance Agentic Search},
  author={Li, Zhanli and Tian, Huiwen and Luo, Lvzhou and Cao, Yixuan and Luo, Ping},
  journal={arXiv preprint arXiv:2602.05014},
  year={2026}
}

@article{liu2025resolving,
  title={Resolving evidence sparsity: Agentic context engineering for long-document understanding},
  author={Liu, Keliang and Chen, Zizhi and Li, Mingcheng and Tang, Jingqun and Yang, Dingkang and Zhang, Lihua},
  journal={arXiv preprint arXiv:2511.22850},
  year={2025}
}

@article{ding2025survey,
  title={A Survey on MLLM-based Visually Rich Document Understanding: Methods, Challenges, and Emerging Trends},
  author={Ding, Yihao and Luo, Siwen and Dai, Yue and Jiang, Yanbei and Li, Zechuan and Martin, Geoffrey and Peng, Yifan},
  journal={arXiv preprint arXiv:2507.09861},
  year={2025}
}

@article{gao2023retrieval,
  title={Retrieval-augmented generation for large language models: A survey},
  author={Gao, Yunfan and Xiong, Yun and Gao, Xinyu and Jia, Kangxiang and Pan, Jinliu and Bi, Yuxi and Dai, Yixin and Sun, Jiawei and Wang, Haofen and Wang, Haofen and others},
  journal={arXiv preprint arXiv:2312.10997},
  volume={2},
  number={1},
  pages={32},
  year={2023}
}

@inproceedings{cuconasu2024power,
  title={The power of noise: Redefining retrieval for rag systems},
  author={Cuconasu, Florin and Trappolini, Giovanni and Siciliano, Federico and Filice, Simone and Campagnano, Cesare and Maarek, Yoelle and Tonellotto, Nicola and Silvestri, Fabrizio},
  booktitle={SIGIR},
  pages={719--729},
  year={2024}
}

@inproceedings{xiong2026docr1,
  title={Docr1: Evidence page-guided grpo for multi-page document understanding},
  author={Xiong, Junyu and Wang, Yonghui and Zhao, Weichao and Liu, Chenyu and Yin, Bing and Zhou, Wengang and Li, Houqiang},
  booktitle={AAAI},
  volume={40},
  number={13},
  pages={11178--11186},
  year={2026}
}

@article{zheng2026doc,
  title={Doc-V*: Coarse-to-Fine Interactive Visual Reasoning for Multi-Page Document VQA},
  author={Zheng, Yuanlei and Fu, Pei and Li, Hang and Wang, Ziyang and Zhang, Yuyi and Ruan, Wenyu and Zhang, Xiaojin and Wei, Zhongyu and Luo, Zhenbo and Luan, Jian and others},
  journal={arXiv preprint arXiv:2604.13731},
  year={2026}
}

@misc{Qwen35,
  author = {{Qwen Team}},
  title = {Qwen3.5: Towards Native Multimodal Agents},
  year = {2026},
  howpublished = {\url{https://qwen.ai/blog?id=qwen3.5}},
}

@inproceedings{napolitano2024leveraging,
  title={On Leveraging Multi-Page Element Relations in Visually-Rich Documents},
  author={Napolitano, Davide and Vaiani, Lorenzo and Cagliero, Luca},
  booktitle={COMPSAC},
  pages={360--365},
  year={2024},
  organization={IEEE}
}

@misc{gpt54,
  author = {{OpenAI}},
  title = {Introducing GPT‑5.4},
  year = {2026},
  howpublished = {\url{https://openai.com/index/introducing-gpt-5-4/}},
}

@article{lee2026meta,
  title={Meta-Harness: End-to-End Optimization of Model Harnesses},
  author={Lee, Yoonho and Nair, Roshen and Zhang, Qizheng and Lee, Kangwook and Khattab, Omar and Finn, Chelsea},
  journal={arXiv preprint arXiv:2603.28052},
  year={2026}
}

@article{ye2023mplug,
  title={mplug-docowl: Modularized multimodal large language model for document understanding},
  author={Ye, Jiabo and Hu, Anwen and Xu, Haiyang and Ye, Qinghao and Yan, Ming and Dan, Yuhao and Zhao, Chenlin and Xu, Guohai and Li, Chenliang and Tian, Junfeng and others},
  journal={arXiv preprint arXiv:2307.02499},
  year={2023}
}

@inproceedings{hu2025mplug,
  title={mplug-docowl2: High-resolution compressing for ocr-free multi-page document understanding},
  author={Hu, Anwen and Xu, Haiyang and Zhang, Liang and Ye, Jiabo and Yan, Ming and Zhang, Ji and Jin, Qin and Huang, Fei and Zhou, Jingren},
  booktitle={ACL},
  pages={5817--5834},
  year={2025}
}

@article{liu2026textmonkey,
  title={Textmonkey: An ocr-free large multimodal model for understanding document},
  author={Liu, Yuliang and Yang, Biao and Liu, Qiang and Li, Zhang and Ma, Zhiyin and Zhang, Shuo and Bai, Xiang},
  journal={IEEE Transactions on Pattern Analysis and Machine Intelligence},
  year={2026},
}

@inproceedings{deng2025longdocurl,
  title={Longdocurl: a comprehensive multimodal long document benchmark integrating understanding, reasoning, and locating},
  author={Deng, Chao and Yuan, Jiale and Bu, Pi and Wang, Peijie and Li, Zhong-Zhi and Xu, Jian and Li, Xiao-Hui and Gao, Yuan and Song, Jun and Zheng, Bo and others},
  booktitle={ACL},
  pages={1135--1159},
  year={2025}
}

@inproceedings{zhao2025finragbench,
  title={Finragbench-v: A benchmark for multimodal rag with visual citation in the financial domain},
  author={Zhao, Suifeng and Jin, Zhuoran and Li, Sujian and Gao, Jun},
  booktitle={EMNLP},
  pages={4215--4249},
  year={2025}
}

@inproceedings{ma2024mmlongbench,
  author       = {Yubo Ma and
                  Yuhang Zang and
                  Liangyu Chen and
                  Meiqi Chen and
                  Yizhu Jiao and
                  Xinze Li and
                  Xinyuan Lu and
                  Ziyu Liu and
                  Yan Ma and
                  Xiaoyi Dong and
                  Pan Zhang and
                  Liangming Pan and
                  Yu{-}Gang Jiang and
                  Jiaqi Wang and
                  Yixin Cao and
                  Aixin Sun},
  title        = {{MMLONGBENCH-DOC:} Benchmarking Long-context Document Understanding
                  with Visualizations},
  booktitle    = {NeurIPS},
  address = {BC, Canada},
  year         = {2024}
}

@article{liu2024lost,
  author       = {Nelson F. Liu and
                  Kevin Lin and
                  John Hewitt and
                  Ashwin Paranjape and
                  Michele Bevilacqua and
                  Fabio Petroni and
                  Percy Liang},
  title        = {Lost in the Middle: How Language Models Use Long Contexts},
  journal      = {Trans. Assoc. Comput. Linguistics},
  volume       = {12},
  pages        = {157--173},
  year         = {2024}
}

@inproceedings{yang2025docagent,
  title={Docagent: A multi-agent system for automated code documentation generation},
  author={Yang, Dayu and Simoulin, Antoine and Qian, Xin and Liu, Xiaoyi and Cao, Yuwei and Teng, Zhaopu and Yang, Grey},
  booktitle={ACL},
  pages={460--471},
  year={2025}
}

@inproceedings{jain2025simpledoc,
  title={SimpleDoc: Multi-Modal Document Understanding with Dual-Cue Page Retrieval and Iterative Refinement},
  author={Jain, Chelsi and Wu, Yiran and Zeng, Yifan and Liu, Jiale and Dai, Shengyu and Shao, Zhenwen and Wu, Qingyun and Wang, Huazheng},
  booktitle={EMNLP},
  pages={28398--28415},
  year={2025}
}

@article{zhu2025doclens,
  title={Doclens: A tool-augmented multi-agent framework for long visual document understanding},
  author={Zhu, Dawei and Meng, Rui and Chen, Jiefeng and Li, Sujian and Pfister, Tomas and Yoon, Jinsung},
  journal={arXiv preprint arXiv:2511.11552},
  year={2025}
}

@article{zhang2026docdancer,
  title={DocDancer: Towards Agentic Document-Grounded Information Seeking},
  author={Zhang, Qintong and Lv, Xinjie and Wu, Jialong and Li, Baixuan and Tao, Zhengwei and Yan, Guochen and Zhang, Huanyao and Wang, Bin and Xu, Jiahao and Mi, Haitao and others},
  journal={arXiv preprint arXiv:2601.05163},
  year={2026}
}

@article{yu2025mact,
  title={Visual document understanding and reasoning: A multi-agent collaboration framework with agent-wise adaptive test-time scaling},
  author={Yu, Xinlei and Xu, Chengming and Chen, Zhangquan and Zhang, Yudong and Lu, Shilin and Yang, Cheng and Zhang, Jiangning and Yan, Shuicheng and Hu, Xiaobin},
  journal={arXiv preprint arXiv:2508.03404},
  year={2025}
}

@article{yu2025dapo,
  title={Dapo: An open-source llm reinforcement learning system at scale},
  author={Yu, Qiying and Zhang, Zheng and Zhu, Ruofei and Yuan, Yufeng and Zuo, Xiaochen and Yue, Yu and Dai, Weinan and Fan, Tiantian and Liu, Gaohong and Liu, Lingjun and others},
  journal={arXiv preprint arXiv:2503.14476},
  year={2025}
}

@article{shao2024deepseekmath,
  title={Deepseekmath: Pushing the limits of mathematical reasoning in open language models},
  author={Shao, Zhihong and Wang, Peiyi and Zhu, Qihao and Xu, Runxin and Song, Junxiao and Bi, Xiao and Zhang, Haowei and Zhang, Mingchuan and Li, YK and Wu, Yang and others},
  journal={arXiv preprint arXiv:2402.03300},
  year={2024}
}

@inproceedings{niu2025mineru2,
  title={Mineru2.5: A decoupled vision-language model for efficient high-resolution document parsing},
  author={Niu, Junbo and Liu, Zheng and Gu, Zhuangcheng and Wang, Bin and Ouyang, Linke and Zhao, Zhiyuan and Chu, Tao and He, Tianyao and Wu, Fan and Zhang, Qintong and others},
  booktitle={ACL},
  year={2025}
}

@article{han2025mdocagent,
  title={Mdocagent: A multi-modal multi-agent framework for document understanding},
  author={Han, Siwei and Xia, Peng and Zhang, Ruiyi and Sun, Tong and Li, Yun and Zhu, Hongtu and Yao, Huaxiu},
  journal={arXiv preprint arXiv:2503.13964},
  year={2025}
}

@inproceedings{yao2023react,
  author       = {Shunyu Yao and
                  Jeffrey Zhao and
                  Dian Yu and
                  Nan Du and
                  Izhak Shafran and
                  Karthik R. Narasimhan and
                  Yuan Cao},
  title        = {ReAct: Synergizing Reasoning and Acting in Language Models},
  booktitle    = {ICLR},
  address = {Kigali, Rwanda},
  year         = {2023}
}

@inproceedings{sun2025docagent,
  title={Docagent: An agentic framework for multi-modal long-context document understanding},
  author={Sun, Li and He, Liu and Jia, Shuyue and He, Yangfan and You, Chenyu},
  booktitle={EMNLP},
  pages={17712--17727},
  year={2025}
}

@inproceedings{wu2025docreact,
  author       = {Junda Wu and
                  Yu Xia and
                  Tong Yu and
                  Xiang Chen and
                  Sai Sree Harsha and
                  Akash V. Maharaj and
                  Ruiyi Zhang and
                  Victor S. Bursztyn and
                  Sungchul Kim and
                  Ryan A. Rossi and
                  Julian J. McAuley and
                  Yunyao Li and
                  Ritwik Sinha},
  title        = {Doc-React: Multi-page Heterogeneous Document Question-answering},
  booktitle    = {ACL},
  pages        = {67--78},
  address    = {Vienna, Austria},
  year         = {2025}
}

@inproceedings{FaysseSWOVHC25,
  author       = {Manuel Faysse and
                  Hugues Sibille and
                  Tony Wu and
                  Bilel Omrani and
                  Gautier Viaud and
                  C{\'{e}}line Hudelot and
                  Pierre Colombo},
  title        = {ColPali: Efficient Document Retrieval with Vision Language Models},
  booktitle    = {ICLR},
  year         = {2025}
}

@inproceedings{ma2023query,
  title={Query rewriting in retrieval-augmented large language models},
  author={Ma, Xinbei and Gong, Yeyun and He, Pengcheng and Zhao, Hai and Duan, Nan},
  booktitle={EMNLP},
  pages={5303--5315},
  year={2023}
}

@inproceedings{asai2024selfrag,
  title={Self-rag: Learning to retrieve, generate, and critique through self-reflection},
  author={Asai, Akari and Wu, Zeqiu and Wang, Yizhong and Sil, Avirup and Hajishirzi, Hannaneh},
  booktitle={ICLR},
  year={2023}
}

@article{lou2026autoharness,
  title={AutoHarness: improving LLM agents by automatically synthesizing a code harness},
  author={Lou, Xinghua and L{\'a}zaro-Gredilla, Miguel and Dedieu, Antoine and Wendelken, Carter and Lehrach, Wolfgang and Murphy, Kevin P},
  journal={arXiv preprint arXiv:2603.03329},
  year={2026}
}

@article{yu2024visrag,
  title={VisRAG: Vision-based Retrieval-augmented Generation on Multi-modality Documents},
  author={Yu, Shi and Tang, Chaoyue and Xu, Bokai and Cui, Junbo and Ran, Junhao and Yan, Yukun and Liu, Zhenghao and Wang, Shuo and Han, Xu and Liu, Zhiyuan and Sun, Maosong},
  journal={arXiv preprint arXiv:2410.10594},
  year={2024}
}

@article{mace2025vidore,
  title={ViDoRe Benchmark V2: Raising the Bar for Visual Retrieval},
  author={Mac{\'e}, Quentin and Loison, Ant{\^o}nio and Faysse, Manuel},
  journal={arXiv preprint arXiv:2505.17166},
  year={2025}
}

@article{osmulski2025miraclvision,
  title={MIRACL-VISION: A Large, Multilingual, Visual Document Retrieval Benchmark},
  author={Osmulski, Radek and de Souza P. Moreira, Gabriel and Ak, Ronay and Xu, Mengyao and Schifferer, Benedikt and Oldridge, Even},
  journal={arXiv preprint arXiv:2505.11651},
  year={2025}
}
